%% file: melting_pot_competition_data.tex
\documentclass{ecai} 

\usepackage{latexsym}
\usepackage{amssymb}
\usepackage{amsmath}
\usepackage{amsthm}
\usepackage{booktabs}
\usepackage{enumitem}
\usepackage{graphicx}
\usepackage{color}
\usepackage{longtable}
\usepackage{bm}

\newcommand{\BibTeX}{B\kern-.05em{\sc i\kern-.025em b}\kern-.08em\TeX}

\begin{document}


\begin{frontmatter}


\paperid{6886} 


\title{Beyond the high score: Prosocial ability profiles of multi-agent populations}


\author[A]{\fnms{Marko}~\snm{Tešić}\thanks{Corresponding Author. Email: mt961@cam.ac.uk}} 
\author[B]{\fnms{Yue}~\snm{Zhao}}
\author[C]{\fnms{Joel}~\snm{Z. Leibo}}
\author[D]{\fnms{Rakshit}~\snm{S. Trivedi}}
\author[A,E]{\fnms{José}~\snm{Hernández-Orallo}}

\address[A]{University of Cambridge}
\address[B]{Northwestern Polytechnical University}
\address[C]{Google DeepMind}
\address[D]{MIT}
\address[E]{VRAIN, Universitat Politècnica de València}


\begin{abstract}
The development and evaluation of social capabilities in AI agents require complex environments where competitive and cooperative behaviours naturally emerge. While game-theoretic properties can explain why certain teams or agent populations outperform others, more abstract behaviours, such as convention following, are harder to control in training and evaluation settings. The Melting Pot contest is a social AI evaluation suite designed to assess the cooperation capabilities of AI systems.
In this paper, we apply a Bayesian approach known as Measurement Layouts to infer the capability profiles of multi-agent systems in the Melting Pot contest. We show that these capability profiles not only predict future performance within the Melting Pot suite but also reveal the underlying prosocial abilities of agents. Our analysis indicates that while higher prosocial capabilities sometimes correlate with better performance, this is not a universal trend—some lower-scoring agents exhibit stronger cooperation abilities. Furthermore, we find that top-performing contest submissions are more likely to achieve high scores in scenarios where prosocial capabilities are not required. These findings, together with reports that the contest winner used a hard-coded solution tailored to specific environments, suggest that at least one top-performing team may have optimised for conditions where cooperation was not necessary, potentially exploiting limitations in the evaluation framework.
We provide recommendations for improving the annotation of cooperation demands and propose future research directions to account for biases introduced by different testing environments. Our results demonstrate that Measurement Layouts offer both strong predictive accuracy and actionable insights, contributing to a more transparent and generalisable approach to evaluating AI systems in complex social settings.
\end{abstract}

\end{frontmatter}


\section{Introduction}\label{sec:Intr}

As AI systems become more widespread and capable, the need for robust evaluation grows increasingly urgent. Simple aggregate metrics fail to capture the full range and complexity of their capabilities. A more comprehensive and nuanced evaluation approach is needed.

Analysing capabilities is a crucial step in conducting detailed evaluations \cite{Burnell}. A capability profile---representing the levels of various capabilities across multiple dimensions---helps identify a system’s strengths and weaknesses, guiding improvements and mitigating failures. A robust approach to achieving more granular capability analysis is Measurement Layouts \cite{burden2023inferring}, a hierarchical Bayesian network framework that enables both explanatory and predictive modelling of AI systems behaviour. In backward inference, measurement layouts infer a system’s capability and bias profile based on task demands and observed performance, allowing us to explain an AI system’s behaviour in terms of their capability profile. In forward inference, they predict performance on new tasks by leveraging task demands and the system’s inferred capability profile. By applying measurement layouts, we can make reliable performance predictions for unseen tasks, diagnose the causes of AI system failures, and precisely characterise the strengths and limitations of different AI architectures. This method provides a robust inference framework applicable to a wide range of AI systems, including reinforcement learning agents and language models \cite{rutar2024general,mlayoutstutorial2024}.

In this paper, we apply Measurement Layouts to a multi-agent context. Specifically, we analyse agents tested in Melting Pot, a scalable multi-agent evaluation platform designed to assess cooperative intelligence \cite{agapiou2023melting,leibo2021meltingpot}. We demonstrate how Measurement Layouts can be used to infer the prosocial abilities of agents that participated in the Melting Pot Contest \cite{trivedi2025melting}. This contest leverages the Melting Pot framework to systematically evaluate how well multi-agent algorithms exhibit cooperative behaviour when interacting with diverse agent populations across various environments. 

In our previous work, we used Measurement Layouts to explore a broad range of factors—including the environments in which agents were tested and their roles in those environments (i.e., whether they were visitors or residents)—that contribute to their performance in the Melting Pot Contest \cite{trivedi2025melting}. In this study, we specifically focus on prosocial abilities, which enable agents to perform well in cooperative multi-agent settings. Furthermore, we compare the predictive power of Measurement Layouts to the assessor approach, a black-box model used to predict AI system performance \cite{hernandez2022training}.

Our analysis shows that the capability profiles generated by Measurement Layouts provide detailed insights into agent behaviour. While some prosocial abilities correlate with higher performance, this relationship is not consistently observed and some lower-scoring agents exhibit stronger performance in specific cooperation dimensions. Notably, we found that top-performing entries were more likely to achieve higher scores when prosocial demands were absent (i.e.~they had higher base chance rates). The competition authors \cite{trivedi2025melting} report that the competition winner implemented hard-coded policy tailored to specific substrates. Combined with our findings, this suggests that certain teams, whether they submitted hard-coded strategies or learning-based policies, may have targeted scenarios in which cooperation conferred little advantage, thereby securing higher scores by exploiting gaps in the evaluation framework. To improve future evaluations, we recommend refining demand annotations to ensure more comprehensive and non-redundant task characterisation while reducing sparsity. Additionally, future research should explore modelling substrates as bias factors that influence capability expression, as environmental constraints can significantly affect agent performance.


\section{Background}\label{sec:Bac}

Traditionally, AI model evaluation focuses on measuring performance against standardised benchmarks or test datasets, primarily assessing whether a system can successfully complete predefined tasks. However, as AI systems become more general, it is increasingly important to evaluate their ability to handle novel tasks, particularly those that cannot be anticipated from the training data. One approach to addressing this challenge is capability-oriented evaluation \cite{hernandez2017evaluation}. Capability-oriented evaluation of AI systems involves a wide range of methodologies, which can be broadly classified into two major categories:~(1) alternative uses of benchmarks, where existing datasets are leveraged in new ways to extract latent factors representing capabilities and (2) approaches inspired by psychology and cognitive science, where prior domain knowledge about how cognitive processes or demands affect performance can be use to model capabilities. 

In either case, estimating capabilities requires a more granular analysis of system performance---instance-level evaluation, where individual data examples are taken separately. This could involve breaking down results according to features of the problem space that are either hypothesised or empirically shown to affect performance, as suggested by \cite{Burnell}. For example, tasks can be designed to assess specific concepts or cognitive capabilities while ensuring that task instances vary systematically along important dimensions of the problem space. 

This approach contrasts with the common practice of aggregating scores, which makes it difficult to derive nuanced insights about system failures and their underlying causes at the instance level. While high-level summary statistics can provide a general sense of a system's capabilities, they often obscure crucial details and fail to offer a comprehensive understanding of abstract and latent capabilities. Instead, performance breakdowns should be guided by relevant problem-space features that influence performance, either replacing or complementing aggregate metrics. However, partitioning a dataset into $N$ domains and calculating partial performance presents several challenges. Firstly, the performance in each domain depends on the difficulty of instances within that partition. For example, a system achieving 30\% in Domain A and 80\% in Domain B cannot be conclusively deemed better at Domain B if the domains are different in difficulty. Secondly, these partial performance measures are often not generalisable to new situations:~performance extrapolation is only valid for datasets with a same partition of domains and {\em same difficulty levels in each domain}. To address these limitations, new families of AI evaluation have been introduced. One approach, assessors, uses black-box models to predict system performance. Another, Measurement Layout, takes a cognitively inspired approach, predicting and \textit{explaining} performance through the inference of capability profiles. In the following sections, we explore each of these families before returning to the problem of evaluating cooperative intelligence.

\subsection{Assessors}

Predicting performance or other indicators of an AI system can be effectively achieved by training separate prediction models using {\em test data} \cite{hernandez2022training}. A straightforward approach involves collecting test data from various system and task instances and using it to train an ``assessor'' model. These models aim to establish a general mapping between the space of AI systems, the space of task instances, and the corresponding distribution of performance scores. An assessor is expressed as follows:

\begin{equation}\label{accss}
  \hat{R}(r|\pi,\mu)~\approx~\Pr(R(\pi,\mu)=r).  
\end{equation}

Here, $\pi$ represents a profile or description of a given system, while $\mu$ denotes a specific problem instance or task example. $R (\pi, \mu) $ is the score function. The measurements $r$ are outcomes of $R (\pi, \mu) $ and follow the conditional distribution ${R} (r | \pi, \mu) $. The assessor model estimates this distribution, predicting the likelihood of different values of $r$ for a given system $\pi$ and task instance $\mu$.

The construction, interpretation, and evaluation of an assessor depend on the definition of its input and output spaces. For instance, if we have identified the features of the problem that might be predictive of performance, we could simply use standard feature-based machine learning methods such as logistic regression or XGBoost to train the assessor. Once trained, the assessor will output predictions that can be evaluated using traditional loss metrics. For instance, if the assessor's output is a binary decision indicating whether the AI system's response to a given test instance will be correct or incorrect, then the assessor is solving a binary probabilistic prediction problem, estimating the probability of correctness. In this case, evaluation metrics such as the area under the ROC curve (AUC) or the Brier score can be used to evaluate the assessor's predictive power. If the goal is instead to estimate the AI system's score on a continuous scale, then the assessor is solving a regression problem, where performance can be evaluated using metrics such as the coefficient of determination ($R^2$) and mean squared error (MSE).

A key limitation of assessor models is that the most effective training algorithms, such as XGBoost, do not inherently provide insights into how different features influence outcomes. Even logistic regression, which produces feature coefficients, fails to capture the notion of capability. A capability is an intrinsic property of a system and should remain unchanged regardless of the distribution of tasks or their difficulty levels in the evaluation set. However, the learned coefficients in models like logistic regression are inherently dependent on the specific distribution of tasks the system has been tested on. If the distribution of test instances were different, the coefficients would also change. As a result, these coefficients may not accurately reflect the system's true capabilities.

Nonlinear models introduce additional challenges. If an assessor learns a nonmonotonic relationship---where success occurs at both low and high demand levels but not at intermediate levels---this suggests an unrealistic pattern that does not align with how capabilities often function. These limitations make it difficult to accurately model agent behaviour and assess their true constraints. In this paper, we build assessors to predict the performance of submissions to the Melting Pot contest and use them as a baseline for comparing the predictive power of measurement layouts.

\subsection{Measurement Layout}

Measurement Layouts \cite{burden2023inferring} are Bayesian models designed to infer an agent's capability profile and predict performance based on that profile. These models use an agent’s observed performance and the task-instance demands (meta-features) to estimate its latent capabilities. Once inferred, these capabilities are then used in conjunction with the demands of new tasks to predict the agent’s performance on those tasks. In general, a Measurement Layout is represented as a directed acyclic graph (DAG), where meta-features of a task (i.e.~task demands) are linked to a system's performance on that task. The structure consists of three key components: root nodes, derived nodes, and linking functions.
\begin{itemize}
\item  Root nodes. The root nodes of a Measurement Layout include the meta-features of the task instance (the demands) and the cognitive profile of the system (its capabilities, noise, and biases). The distribution and range of these elements are determined by the specific characteristics of the task.

\item Derived nodes. These nodes are inferred from their dependent root nodes and include both child nodes and leaf nodes. Child nodes represent unobservable intermediate states, such as partial or intermediate performance based on a specific capability and task demand. Leaf nodes, on the other hand, correspond to observable task performance outcomes.

\item Linking functions. A linking function is a mathematical expression that defines how the probability distribution of one node influences the input of another. These functions are often non-linear and capture the hierarchical relationships between abilities and task demands. They also specify how partial performances are combined to generate the final performance scores. Linking functions are developed based on domain expertise related to the task and draw upon concepts from cognitive modelling. 
\end{itemize}

\noindent 
 
Most linking functions that connect capabilities, demands, and performance are closely related to Item Response Theory (IRT) models \cite{embretson1996item}. However, unlike in IRT, both demands and performance are observed in Measurement Layouts. IRT models typically use a logistic function to model the probability of success based on an agent’s capability and the task demand. For a single capability-demand pair, this is expressed as:
\begin{equation}
P(\text{success} \mid \text{capability, demand}) = \sigma(\text{capability}, \text{demand})
\end{equation}
where $\sigma$ is a linking function, such as a logistic function of the difference between capability and demand. The probability of success should increase as capability exceeds task demand. When multiple capabilities and demands are involved, they are connected through partial performance nodes, which capture intermediate contributions to overall performance. The final performance outcome is then derived from an aggregation of these partial performances.

Measurement Layouts can naturally represent tasks that require multiple capabilities, including dependencies between those capabilities. They provide a structured way to infer capability profiles for a wide range of agents, such as reinforcement learning agents, language models, and even humans.

To implement Measurement Layouts, we can use PyMC \cite{salvatier2016probabilistic}, a probabilistic programming framework for Bayesian modelling and inference. Before constructing a model in PyMC, we need to define the meta-features, capabilities, and linking functions so they can be appropriately represented within the Measurement Layout framework. In Section \ref{Sec: measurement layout implementation}, we present a specific implementation of a Measurement Layout designed to infer the capabilities of agents based on their performance and cooperation demands in the Melting Pot contest. PyMC's inference engine primarily relies on the No U-Turn Sampler (NUTS), an adaptive extension of Hamiltonian Monte Carlo (HMC). NUTS automatically tunes step sizes and trajectory lengths, eliminating the need for manual hyperparameter selection, making it well-suited for complex hierarchical models such as those used in Measurement Layouts.

\subsection{Competition and Cooperation Traits in Social AI}

In social AI, interactions between individual agents or teams can give rise to emergent behaviours, which may be either cooperative or competitive \cite{castelfranchi1998modelling}. As AI systems continue to advance, multi-agent systems are becoming more autonomous and efficient, achieving higher performance in both cooperative and adversarial settings while adapting more effectively to complex environments. Evaluating the performance of individual agents or teams is essential for understanding their behaviour and uncovering the mechanisms that drive cooperation or competition.

The evaluation of competitive and cooperative games lies at the core of game theory and has applications across many disciplines. In purely cooperative games, players form a coalition, a group that must collaborate to achieve a common goal. When the team is homogeneous, evaluation is straightforward:~the system assesses $n$ identical copies of an algorithm or policy, and the best-performing one can be selected by averaging results. However, in heterogeneous teams, assessing individual contributions becomes more complex due to interdependent interactions---a challenge known as the attribution problem. The Shapley value \cite{roth1988shapley} has emerged as a canonical tool in multi-agent systems for quantifying each agent’s contribution to the overall team performance. Given all possible coalitions and their associated payoffs, the Shapley value assigns a fair contribution score to each player. However, computing the Shapley value is computationally demanding due to the combinatorial explosion in possible team formations. To address this, approximation methods have been developed both to reduce computational costs \cite{fatima2008linear} and, more importantly, to minimise the number of experiments or game simulations required. Despite these challenges, the Shapley value remains a crucial tool in cooperative game theory for analyzing strong interdependencies between players \cite{kotthoff2018quantifying,Shapley2021}. Furthermore, recent work shows that deep reinforcement learning agents trained to negotiate team formation achieved rewards that exhibited high correlation with the Shapley value solution \cite{bachrach2020negotiating}. Extensions of the Shapley value have also been applied to competitive (non-cooperative) games, where it serves as an evaluation metric across varying degrees of heterogeneity—from fully homogeneous teams to entirely heterogeneous teams with no duplicate agents \cite{Heterogeneity}. 

In competitive (non-cooperative) games, ranking methods such as ELO \cite{1978The}, TrueSkill \cite{graepel2007bayesian} and Glicko \cite{glickman1999rating}, primarily evaluate abilities based on pairwise match outcomes. However, in multi-agent system evaluation, many of these methods rely on averaging, which makes them highly sensitive to non-transitivity and the high variability of match results \cite{kiourt2019rating,2019Multiagent}. $\alpha$-Rank addresses these limitations by using inverse reinforcement learning in latent space to train a model capable of predicting game outcomes against different opponents. It is particularly well-suited for complex multi-agent non-cooperative environments, as it adapts to varying numbers and types of participants while mitigating issues of asymmetry and non-transitivity. The Nash equilibrium \cite{Nash1950} is a fundamental solution concept in game theory used for strategy evaluation \cite{balduzzi2018re}. It is invariant to redundant tasks and games, making it a robust theoretical tool. However, in multi-agent systems, finding a Nash equilibrium can be computationally difficult or even impossible \cite{aleksandrov2017pure}. New approaches, such as the Nash average method \cite{balduzzi2018re}, introduce the concept of meta-games, where the maximum entropy Nash equilibrium (maxent Nash) is computed using a pairwise win-rate matrix among $N$ agents. While Nash equilibria provide useful strategic insights, they have high computational complexity, are static, and only describe a specific combination of strategies. 

However, most existing evaluation techniques are limited when assessing team-vs-team interactions, especially in complex social dynamics such as deception. Melting Pot \cite{agapiou2023melting,leibo2021meltingpot} is a multi-agent platform and social AI evaluation suite designed to assess agents and populations in social environments with substantial spatial and temporal complexity. It specifically evaluates generalisation to novel social settings, including interactions with both familiar and unfamiliar individuals. Melting Pot is built to test a broad range of social behaviours, such as collaboration, competition, deception, reciprocation, trust, and stubbornness. The package includes over 50 multi-agent reinforcement learning (MARL) substrates for training agents, along with more than 250 distinct test scenarios for evaluation. An agent’s performance in these test scenarios quantifies its ability to adapt to diverse social situations where individuals are interdependent.

In Melting Pot, a scenario is a multi-agent environment that combines a physical environment (a ``substrate'') with a reference group of co-players (a ``background population'') to create a social setting with significant interdependence between individuals (see Figure \ref{fig:substrates} in Appendix for examples). The substrate refers to the physical environment, which defines the map layout, object placement, movement constraints, physics rules, and other fundamental mechanics. The background population consists of the agents pre-programmed within the simulation, providing a controlled social context. This contrasts with the focal population, which consists of the agents being tested.

Twenty-three teams participated in the Melting Pot contest, with submissions evaluated in three phases. In the final evaluation phase, each team could submit up to three entries, which were tested on 51 held-out scenarios. Each submission was evaluated over 80 episodes, except for 8 lower-performing submissions, which were assessed on 20 episodes.

Melting Pot contest provides a `score' representing the per capita return of the focal population. To ensure consistency across different substrates, these scores are normalised using the Melting Pot 2.0 baseline range \cite{trivedi2025melting}. A normalised score above 1 indicates that the focal agents outperformed the best baseline, while a score below 0 signifies that they performed worse than the worst baseline.

\section{Capability Profiles for Melting Pot Contest}

In this section, we construct capability profiles related to cooperative behaviour for Melting Pot submissions. The section is structured as follows. First, we present the score data collected during the Melting Pot competition. Next, we describe the meta-features (or demands) associated with cooperation capabilities. Finally, using performance scores and demands, we design measurement layouts to build capability profiles and compare their predictive power against various baselines, including the assessors. All code and data are available at \url{https://github.com/markotesic/melting-pot-competition-analysis}.

\subsection{Performance scores}

The Melting Pot competition collected per capita performance scores for the focal population of each submission, tested on 51 held-out scenarios over 80 episodes (or 20 episodes for lower-scoring agents). These scores were normalised using the Melting Pot 2.0 baseline range. For our analysis, rather than averaging scores over all 80 episodes, we preserve more granular information by computing an average score every 5 episodes. A mixed‑effects regression of normalised scores on episode number with random intercepts for each entry yielded a slope of zero (with $z = - 0.203$ and $p = 0.839$; see full results in Table \ref{tab:mixedlm_results} in the Appendix), indicating no systematic episode‑by‑episode trend. Thus, aggregating into five‑episode bins reduces noise without obscuring any effects found in data. This results in 16 performance score data points per scenario (or 4 data points for lower-performing entries). The total dataset size is then 51 scenarios $\times$ 16 (or 4) score data points resulting in 816 (204) rows. The dataset was then split into training and test sets using an 80/20 split, yielding 653 rows in the training set (163 for lower-scoring entries) and 163 rows in the test set (41 for lower-scoring entries). Figure \ref{fig:every_5_episodes} in the Appendix shows the scores for each entry, averaged over every five episodes.

These scores were then Min-Max normalised for use in the measurement layout, where they were modeled using a Beta distribution (see Section~\ref{Sec: measurement layout implementation}). The average scores across all episodes and scenarios are shown in Figure~\ref{fig:average_scores}. The distribution of these scores is presented in Figure~\ref{fig:scores_dist} in Appendix, which also contains other descriptive statistics from the Melting Pot competition.

\begin{figure*}[ht]
    \centering
    \includegraphics[width=0.8\linewidth]{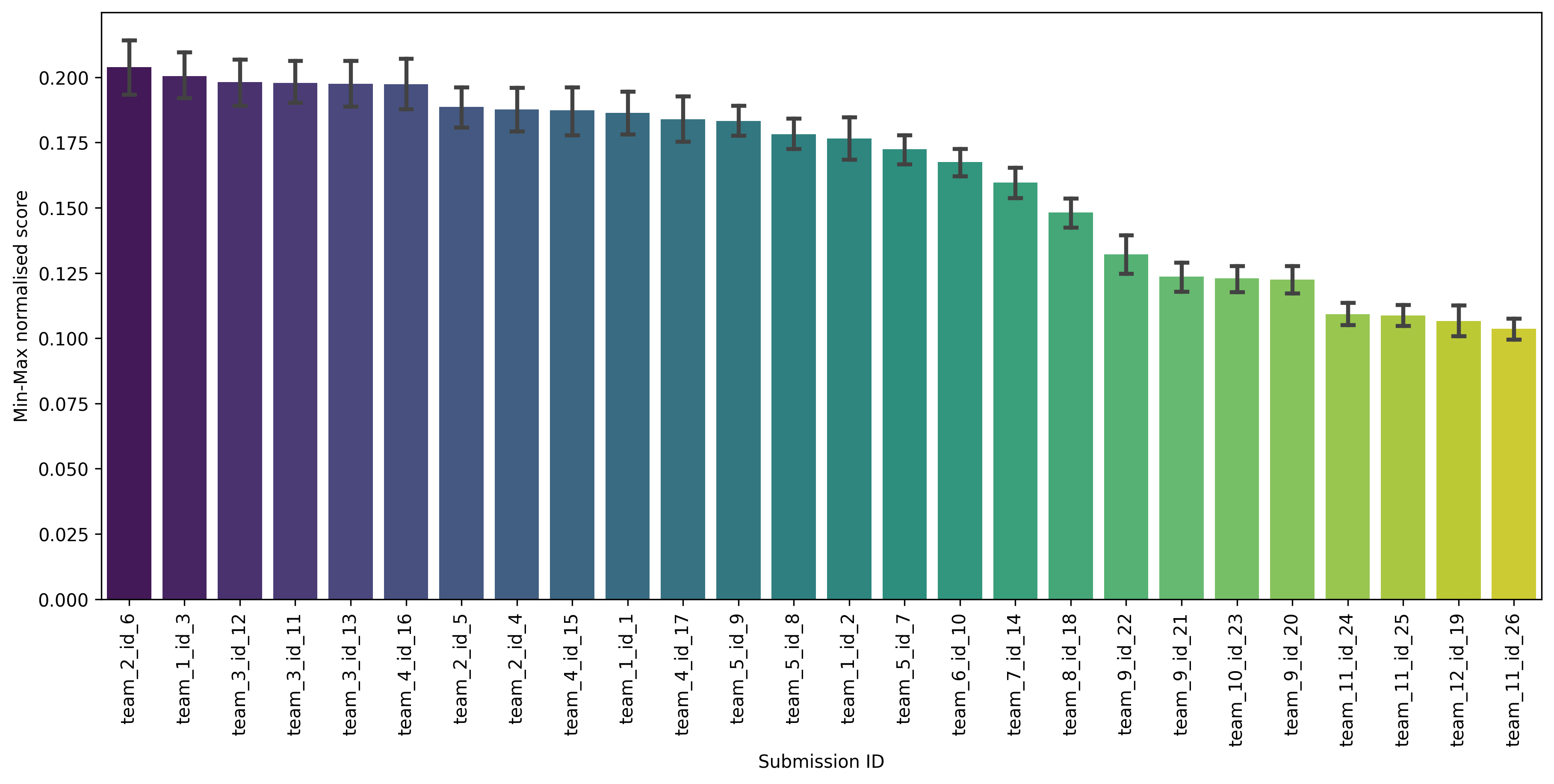}
    \caption{Mean scores and 95\% confidence intervals for Min-Max normalised scores for each submission.}
    \label{fig:average_scores}
\end{figure*}

\subsection{Cooperation demands}

Trivedi et al. \cite{trivedi2025melting} annotated the scenarios using 16 meta-features that describe the demands related to cooperation capabilities tested in the competition. Each demand is binary, meaning it is either present or absent. These demands indicate whether a particular cooperation capability is required in each of the 51 scenarios used to evaluate competition entries. The presence or absence of all 16 demands across the 51 scenarios is visualised in Figure \ref{fig:tags_by_scenario} in the Appendix. We find that some demands are identical, as they are present or absent in the same scenarios: partner\_choice, time\_pressure, and ostracism share identical patterns, as do teaching and sanctioning. To reduce redundancy, we will retain time\_pressure and teaching, removing the duplicated demands.

A correlation matrix for all demands and the Min-Max normalised score confirms these patterns; see Figure \ref{fig:corr_matrix} in the Appendix. Additionally, we observe that most correlations between demands are low, indicating that they capture distinct aspects of cooperative behaviour.

\begin{figure*}
    \centering
    \includegraphics[width=0.9\linewidth]{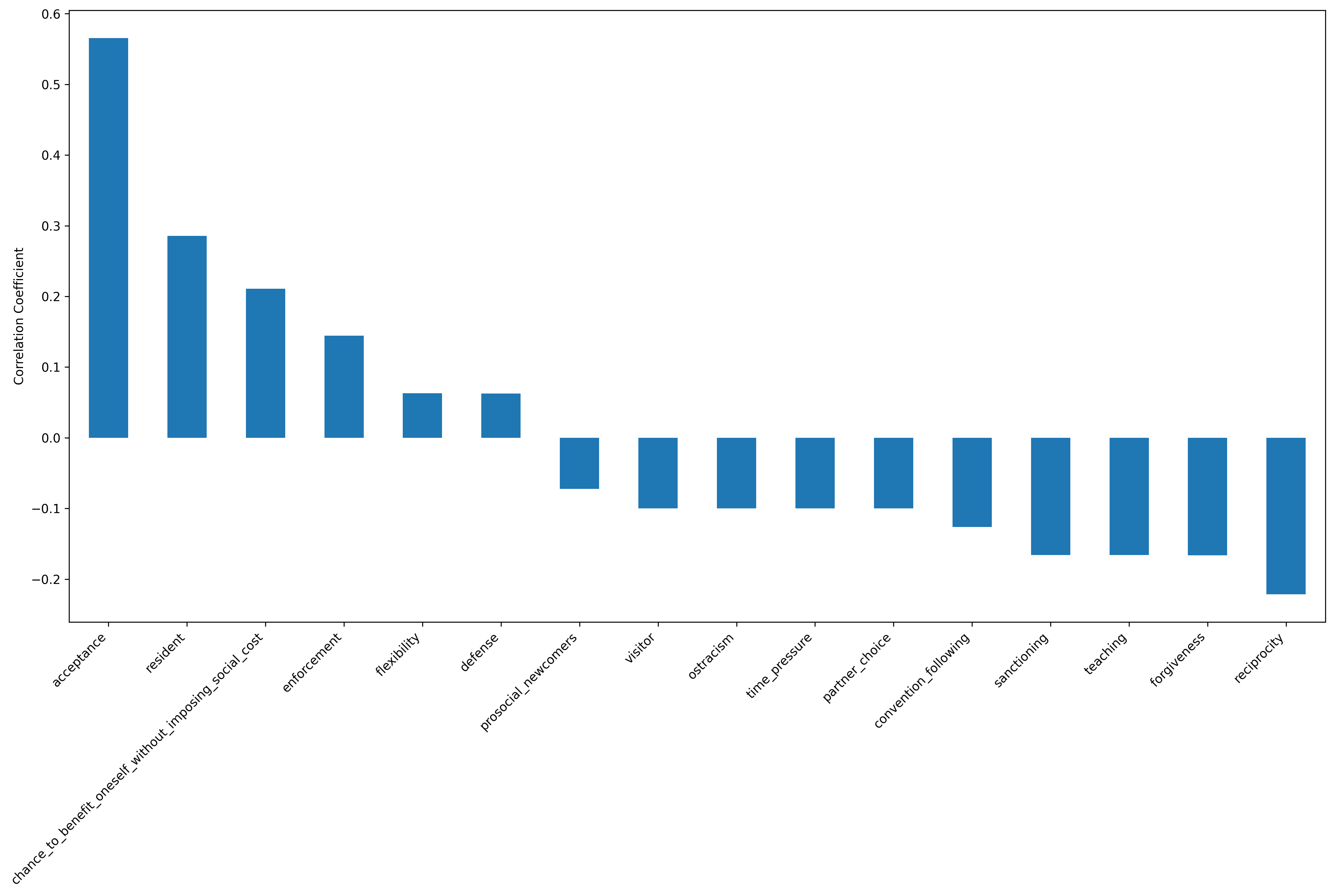}
    \caption{Correlations between the 16 demands and the Min-Max normalised score across all top-scoring submissions, i.e.~those tested for 80 episodes per scenario.}
    \label{fig:corr_with_score}
\end{figure*}

Demands are conceptualised as features of a task that a subject must meet to perform well. In this sense, demands function as determinants of task difficulty, where more challenging tasks require higher levels of capability. Given this, we expect performance to be negatively correlated with demand levels: the higher the demand, the lower the performance score (assuming the subject’s capability remains constant during testing). In the case of binary demands, this implies that a subject must possess the corresponding capability when a specific task demand is present to succeed. Conversely, when the demand is absent, the subject succeeds on the tasks with same base chance (more on on base chance in Section~\ref{Sec: measurement layout implementation}). Consequently, we expect that performance scores will be lower when a demand is present compared to when it is absent. Figure~\ref{fig:corr_with_score} shows the correlation of all demands with the Min-Max normalised score. For further capability analysis, we retain only the demands that are negatively correlated with performance while also removing duplicate demands. This results in seven remaining demands: ‘prosocial\_newcomers’, ‘visitor’, ‘time\_pressure’, ‘convention\_following’, ‘teaching’, ‘forgiveness’, and ‘reciprocity’.


\subsection{Measurement Layout Model for Inferring Cooperation Capabilities}\label{Sec: measurement layout implementation}

\subsubsection{The measurement layout}

Here we present a measurement layout model to infer the cooperation capabilities of submissions to the Melting Pot competition. To achieve this, we use two key pieces of information:~the demands of each scenario that an entry was tested on and the Min-Max normalised performance scores of the entry. We assume that each demand corresponds to a latent capability that we aim to infer. Since we identified seven relevant demands (i.e., those negatively correlated with performance), this means we infer seven capabilities:~abilityTimePressure, abilityReciprocity, abilityProsocialNewcomers, abilityVisitor, abilityForgiveness, abilityConventionFollowing and abilityTeaching. Because demands are binary (present or absent), we infer the probability that a given submission possesses each capability. Specifically, we estimate the likelihood that an entry has the required capability when a corresponding demand is present. If a submission consistently achieves high scores in scenarios where a particular demand is present, it is more likely to possess the associated capability. Conversely, if performance drops significantly when the demand is present, the entry is less likely to have that capability.

To infer the cooperation capabilities of submitted entries, we construct measurement layout, i.e. a hierarchical Bayesian model, that estimates the Min-Max normalised score in different scenarios based on the demands they impose. The model is formalised as follows:

\begin{equation}
    \begin{aligned}
        \theta_j &\sim \mathcal{B}(1, 1), & \text{(Abilities)} \\
        \rho &\sim \mathcal{B}(5, 1), & \text{(Base Chance)} \\
        \lambda_{i,j} &= (1 - \delta_{i,j})\, \rho + \delta_{i,j}\, \theta_j, & \text{(Local Performance)} \\
        \Lambda_i &= \prod_{j} \lambda_{i,j}, & \text{(Integrated Performance)} \\
        \nu &\sim \mathcal{HN}(5), & \text{(`Sample Size')} \\
        p_i &\sim \mathcal{B} \left( \Lambda_i\, \nu, (1 - \Lambda_i)\, \nu \right), & \text{(Observed Score)}
    \end{aligned}
    \label{eq:bayesian_model}
\end{equation}

Each capability, $\theta_j$, is a latent variable representing an entry’s ability to meet demand $j$. It is assigned a non-informative uniform prior. We also introduce a base chance parameter, $\rho$, which represents an entry’s probability of success when no specific demands are present. This parameter captures the entry’s default performance level on a test instance in the absence of cooperation demands. We assume a skewed prior for $\rho$, reflection our expectation that the likelihood of success is typically on Melting Pot high when no cooperation demands are imposed.

Each scenario is tested over multiple episodes, with scores averaged over blocks of five episodes. Thus, each test instance $i$ effectively corresponds to one such block. Multiple demands $j$ can be present in each test instance $i$. These demands are binary indicators, $\delta_{i,j}$, taking a value of 1 if the demand is present and 0 otherwise. The local performance probability  $\lambda_{i,j}$, which defines the probability of success on demand $j$ in test instance $i$, is given by:~$\lambda_{i,j} = (1 - \delta_{i,j})\, \rho + \delta_{i,j}\, \theta_j$. This formulation ensures that when a demand is present ($\delta_{i,j} = 1$), the probability of success depends on the corresponding ability $\theta_j$. Conversely, when the demand is absent ($\delta_{i,j} = 0$), success is determined by the base chance parameter $\rho$.

To determine an entry’s overall probability of success on test instance $i$, we assume that abilities are non-compensatory and take the product of all local performance probabilities, $\Lambda_i$. In other words, having one cooperation ability does compensate for the absence of another:~failure in meeting any particular demand decreases the overall probability of success. Finally, the observed performance score for test instance $i$, $p_i$, is modeled as a Beta-distributed variable with mean $\Lambda_i$ and `sample size' parameter $\nu$ modeled using a Half-Normal distribution. The parameter $\nu$ represents the effective number of trials and is related to the variance of the Beta distribution. Higher values of $\nu$ correspond to lower variance, meaning observed scores cluster more tightly around $\Lambda_i$.

Using the training data, we fit a Bayesian measurement model using PyMC, a probabilistic programming framework for Bayesian inference. We used MCMC sampling with four chains, each running 4000 samples, with a burn-in period of 1000 steps. Inference was performed using the No-U-Turn Sampler \cite{hoffman2014no}. We fitted a separate measurement layout for each competition entry. A graphical representation of the layout is provided in Figure \ref{fig:mes_layout} in the Appendix.

Convergence was assessed using the Gelman-Rubin statistic ($\hat{R}$) and the effective sample size (ESS) for both bulk and tail distributions, confirming that the chains converged to a common posterior distribution. We also found that for all parameters across all entries the bulk ESS exceeded 7000, while the tail ESS was consistently above 4300, indicating low autocorrelation and stable posterior estimates. Visual inspection of trace plots confirmed that all chains mixed well, with no signs of divergence. Full details regarding $\hat{R}$, ESS bulk, ESS tail as well estimates for the parameters and highest density intervals, can be found in Table~\ref{tab:appendix_posterior_stats} in Appendix.

\subsubsection{Comparison to assessors}

To evaluate the predictive power of measurement layouts, we train three assessor models:~a linear regression model, an XGBoost regressor, and a tabPFN model \cite{hollmann2022tabpfn}. These models use demands as input features and aim to predict performance scores based on those demands. Additionally, we introduce a simple baseline that predicts scores using the mean performance across all test instances in the training set. To compare the performance of the assessors and measurement layouts, we use root mean squared error (RMSE) and the coefficient of determination ($R^2$) as evaluation metrics. A full comparison of RMSE and $R^2$ for every submission is shown in Figure \ref{fig:predictive_power} in the Appendix. We find that both the assessor models and the measurement layouts outperform the mean-score baseline for most submissions, with the notable exceptions of the lower-performing submissions, team\_12\_id\_19 and team\_11\_id\_26. 

We also observe that for several submissions, both the assessor models and the measurement layouts exhibit low predictive power. This finding suggests that, in these cases, the demands may not be sufficiently informative to capture the performance characteristics of the entries, or that additional latent factors significantly influence performance. For the subsequent analysis of ability profiles, we restrict our focus to entries for which the measurement layouts achieve a sufficient level of predictive power, defined as $R^2 > 0.25$.

\subsubsection{Ability profiles}

\begin{figure*}[ht]
\centering
\begin{minipage}{1.1\columnwidth}
  \centering
  \includegraphics[width=\columnwidth]{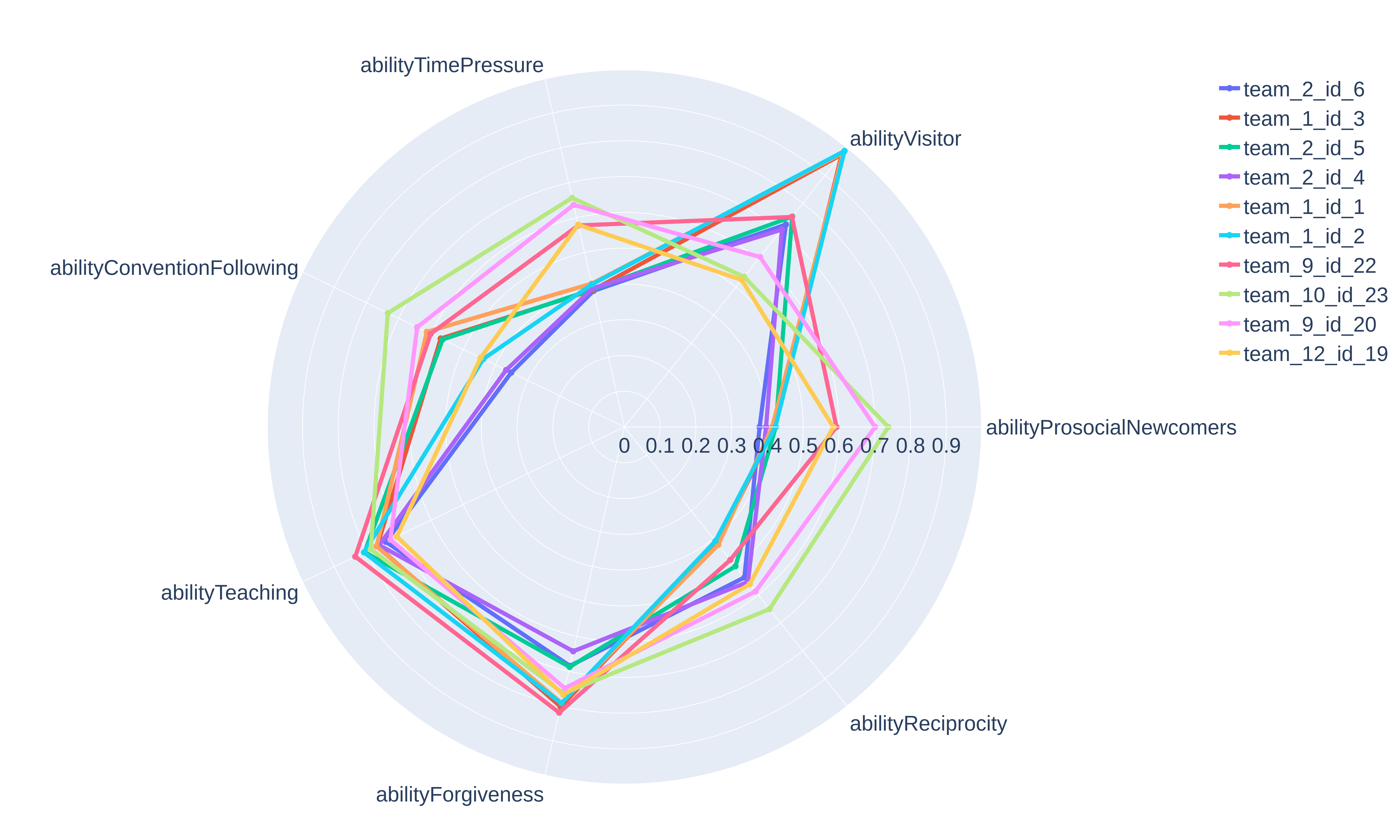}
\end{minipage}\hfill
\begin{minipage}{0.90\columnwidth}
  \centering
  \includegraphics[width=\columnwidth]{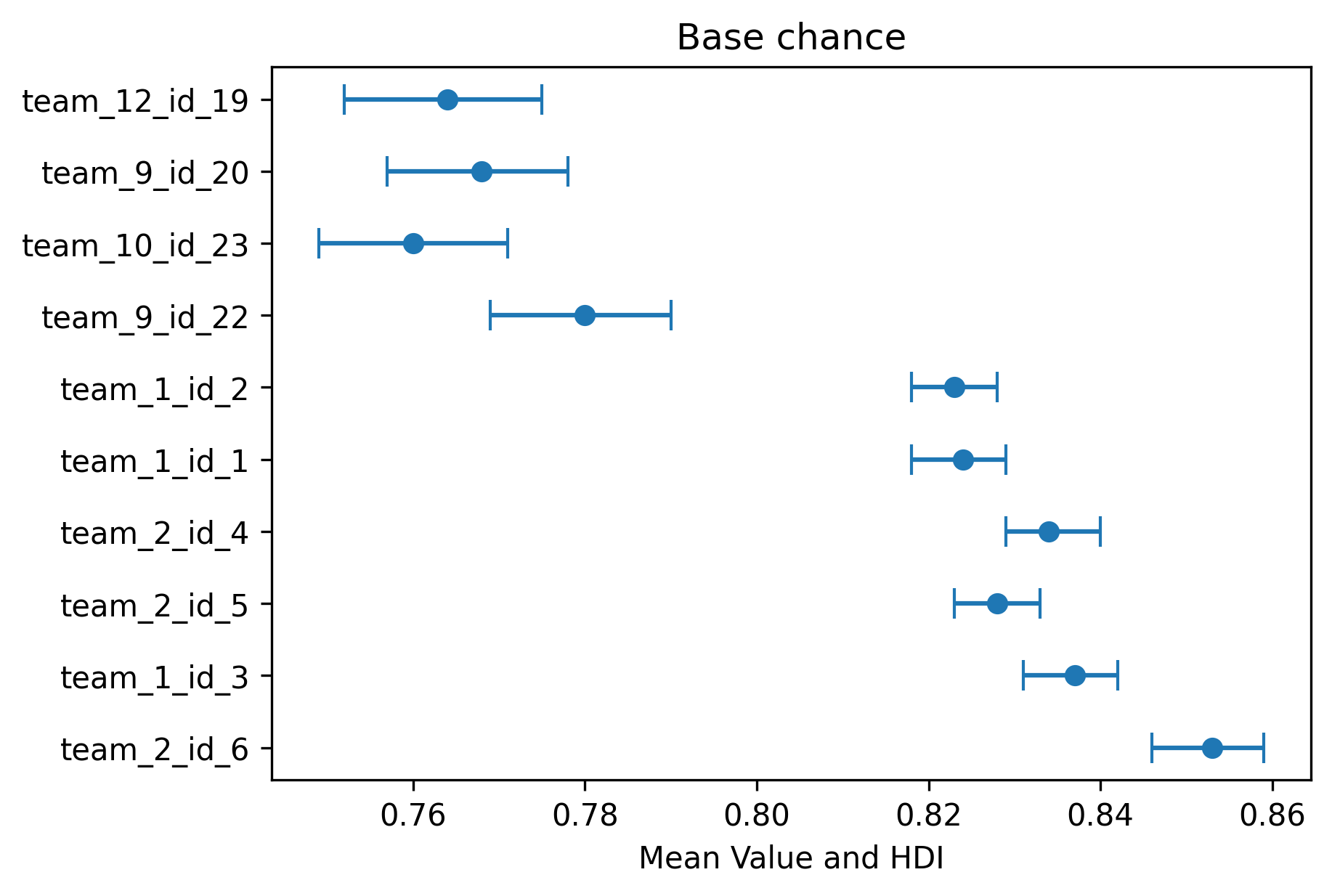}
\end{minipage}
\caption{The left panel displays the estimated ability profiles for competition entries where the measurement layouts had sufficient predictive power. The right panel presents the corresponding base chance values for these entries.}
\label{fig:ability_profile}
\end{figure*}

Figure~\ref{fig:ability_profile} presents the capability profiles for submissions where the measurement layouts demonstrated sufficient predictive power (see Figure \ref{fig:posteriors_hdi} in the Appendix for both the mean estimates and highest density interval (HDI) bounds for posterior estimates of all measurement layout parameters). We observe that higher-scoring submissions (see Figure~\ref{fig:average_scores}) tend to exhibit greater probabilities of possessing certain abilities compared to lower-scoring agents. For example, all three Team 1 submissions achieve higher scores than many lower-scoring agents. However, this pattern is not consistent across all abilities. The results indicate that entries do not significantly differ in their \textit{abilityTeaching}, \textit{abilityForgiveness}, and \textit{abilityTimePressure}. Interestingly, some lower-scoring agents demonstrate higher \textit{abilityProsocialNewcomers} than higher-scoring agents. Additionally, lower-scoring submissions exhibit significantly higher \textit{abilityConventionFollowing} and \textit{abilityReciprocity} compared to some higher-scoring agents. These findings suggest that prosocial abilities alone do not necessarily drive higher scores.

A brief look at the base chance estimates in Figure~\ref{fig:ability_profile} provides further insight into this pattern. The plot shows that higher-scoring entries have significantly higher base chance rates; that is, they are more likely to perform better when prosocial demands are absent. This suggests that some high-scoring agents succeed not because of their prosocial abilities but rather because of general competence or an ability to complete tasks without relying on the specific prosocial capabilities captured by the measurement layouts.


\section{Discussion}\label{sec:discussion}

In this work, we applied Measurement Layouts, a novel evaluation framework, to infer cooperation capability profiles for submissions in the Melting Pot contest. Our approach leverages granular, instance-level performance data and task demands to construct a hierarchical Bayesian model. We demonstrated that Measurement Layouts have predictive power by comparing their performance against traditional assessor models, such as linear regression, XGBoost, and tabPFN. Crucially, Measurement Layouts not only predict performance but also provide ability profiles, allowing us to explain the behaviour of submissions in terms of their prosocial capabilities. Our results indicate that while certain prosocial abilities are linked to higher performance, this is not universally the case—some lower-scoring entries exhibited higher cooperation capabilities than higher-scoring ones. Additionally, some capabilities did not differentiate between submissions, possibly due to the sparse presence of the corresponding demands in the competition.

A notable pattern emerged from the base-chance estimates. Higher-scoring entries exhibit significantly higher base-chance rates, indicating that they perform well even when prosocial demands are absent. The competition report \cite{trivedi2025melting} observes that the single top-ranked submission used a hand-crafted policy tuned to each substrate. Our analysis reveals that several learning-based submissions show similarly high base-chance rates, suggesting that they too might have narrowly specialised for situations in which cooperation conferred no additional advantage. It is then plausible that the leading submissions gained much of their edge by excelling in scenarios that did not reward prosocial ability, thereby exploiting a weakness in the evaluation framework.

Based on our findings, we recommend that future competitions incorporate better-formulated meta-features to characterise task demands while avoiding duplicate demands. Demands should ideally be negatively correlated with performance if they are to serve as meaningful indicators of task difficulty and underlying capabilities. Additionally, the meta-features should be less sparse, allowing for a more robust analysis of capability profiles. Reducing sparsity would increase confidence in inferred capabilities and reduce the risk of competitors exploiting weaknesses in the evaluation framework. Furthermore, although we focused our analysis exclusively on cooperation demands negatively correlated with performance, positively correlated demands might also represent important capabilities if appropriately inverted. However, the current dataset is too sparse to explore this direction effectively, making it a valuable area for future research.

A few limitations are in order. The present demand taxonomy is both sparse and partially redundant. Because only a handful of scenarios exercise some demands, the corresponding ability posteriors may be broad and unable to  distinguish the prosocial abilities between different entries. Also, in this paper we have focused on only one evaluation testbed (Melting Pot). How well the same measurement layout transfers to other social-AI suites remains an open question, although its modular design and generic linking functions should make adaptation feasible, provided that the new scenarios are annotated with an appropriate set of demands. A promising avenue for future work involves modelling the role of substrates. This study did not consider how different substrates might influence an agent’s performance. However, substrates could introduce systematic biases, and certain capabilities may be more or less expressed depending on the substrate an entry is tested on. Future models could treat substrates as bias factors that influence the estimation of latent abilities. While we did not have sufficient data to incorporate this into our current model, extending our framework in this way would provide a more comprehensive understanding of how environmental factors shape cooperation capabilities.

In summary, our work demonstrates that Measurement Layouts offer a dual advantage: they achieve competitive predictive accuracy and explanatory insights into the latent cooperation abilities of agents. This combination is crucial for understanding and improving AI systems in complex social settings, laying the groundwork for more nuanced and interpretable evaluation frameworks in future competitions.






\bibliography{melting_pot_competition_data}
\input{appendix}

\end{document}

%% file: appendix.tex
\clearpage
\appendix
\onecolumn
\section*{Appendix}

\section{Melting Pot competition}

\begin{figure*}[ht]
    \centering
    \includegraphics[width=0.2\textwidth]{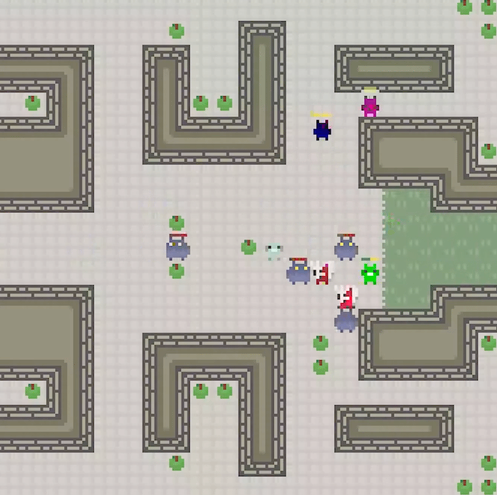}
    \includegraphics[width=0.2\textwidth]{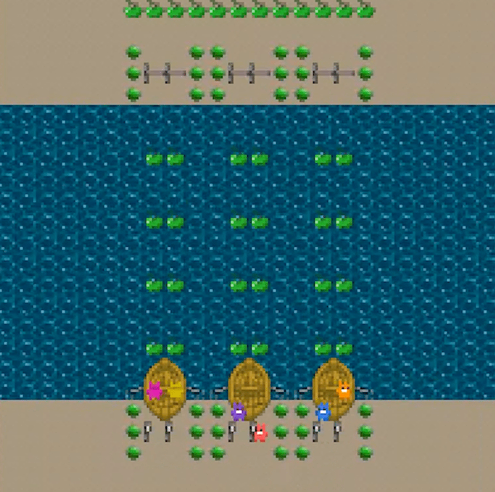}
    \includegraphics[width=0.2\textwidth]{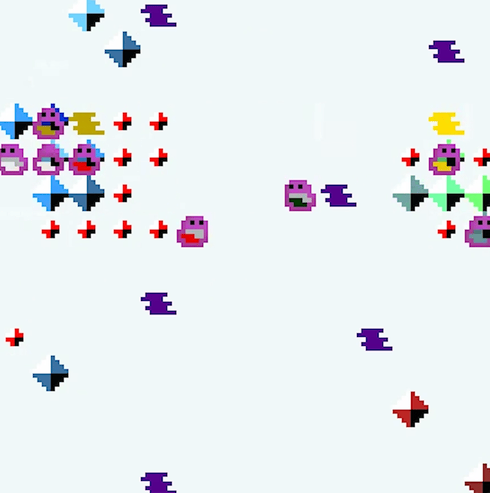}
    \includegraphics[width=0.2\textwidth]{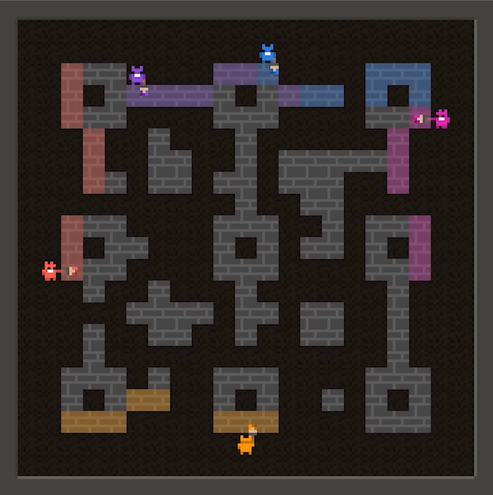}
    
    \includegraphics[width=0.2\textwidth]{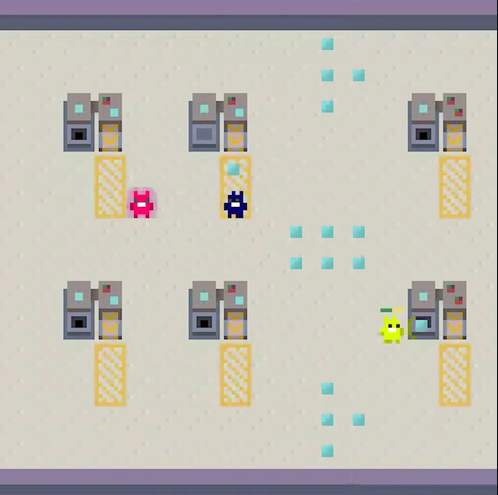}
    \includegraphics[width=0.2\textwidth]{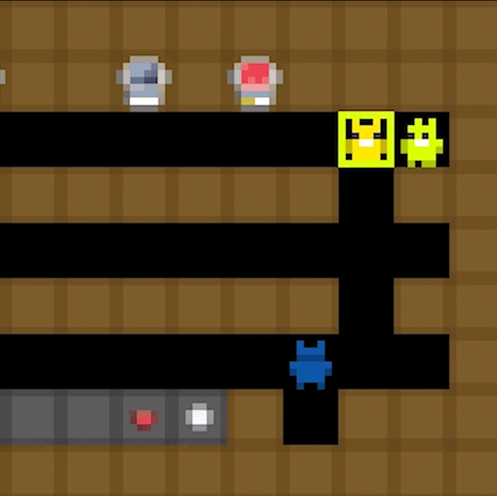}
    \includegraphics[width=0.2\textwidth]{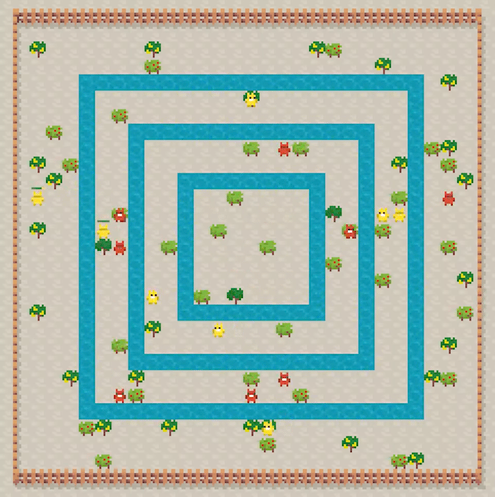}
    \includegraphics[width=0.2\textwidth]{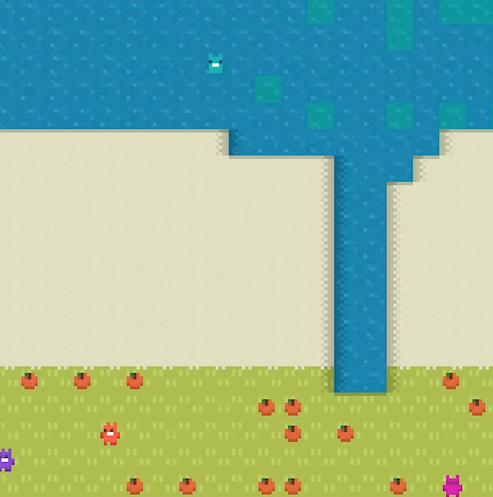}

    \caption{Substrates examples in Melting Pot are 2D physical environments where a background population interact with the world along with the focal population.} \label{fig:substrates}
\end{figure*}

\begin{figure*}[ht]
    \centering
    \includegraphics[width=0.99\linewidth]{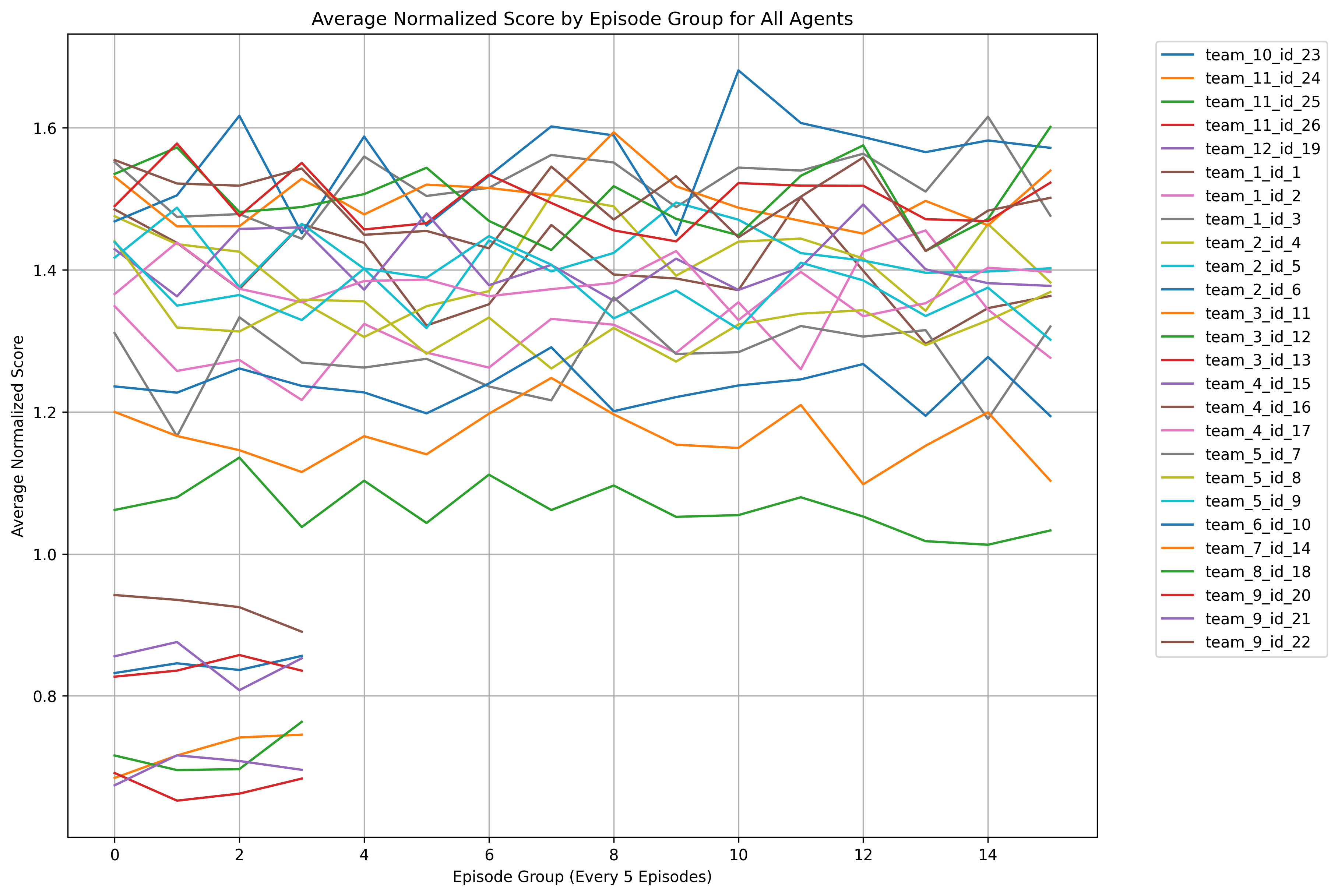}
    \caption{Competition entry performance scores averaged every 5 episodes. The scores in the plot represent the mean performance across all 51 scenarios.} \label{fig:every_5_episodes}
\end{figure*}

\begin{table*}[ht]
\centering
\caption{Mixed‐effects model predicting normalized score from episode number, with a random intercept for each agent.  The model was fitted by REML on 81,600 observations across 26 submissions.}
\label{tab:mixedlm_results}
\begin{tabular}{lrrrrrr}
\toprule
\textbf{Parameter} & \textbf{Estimate} & \textbf{Std.\ Error} & \boldmath{$z$} & \boldmath{$p$} & \textbf{95\% CI lower} & \textbf{95\% CI upper} \\
\midrule
\multicolumn{7}{l}{\itshape Fixed effects}\\
Intercept                  & 1.199 & 0.066 & 18.191 & $<$0.001 & 1.070 & 1.328 \\
Episode No                 & –0.000 & 0.000 & –0.203 & 0.839    & –0.000 & 0.000 \\
\midrule
\multicolumn{7}{l}{\itshape Random effect (variance of intercepts by Submission ID)}\\
Variance (Submission ID)    & 0.112 & 0.034 & —      & —        & —      & —      \\
\bottomrule
\end{tabular}
\end{table*}

\begin{figure*}[ht]
    \centering
    \includegraphics[width=0.9\linewidth]{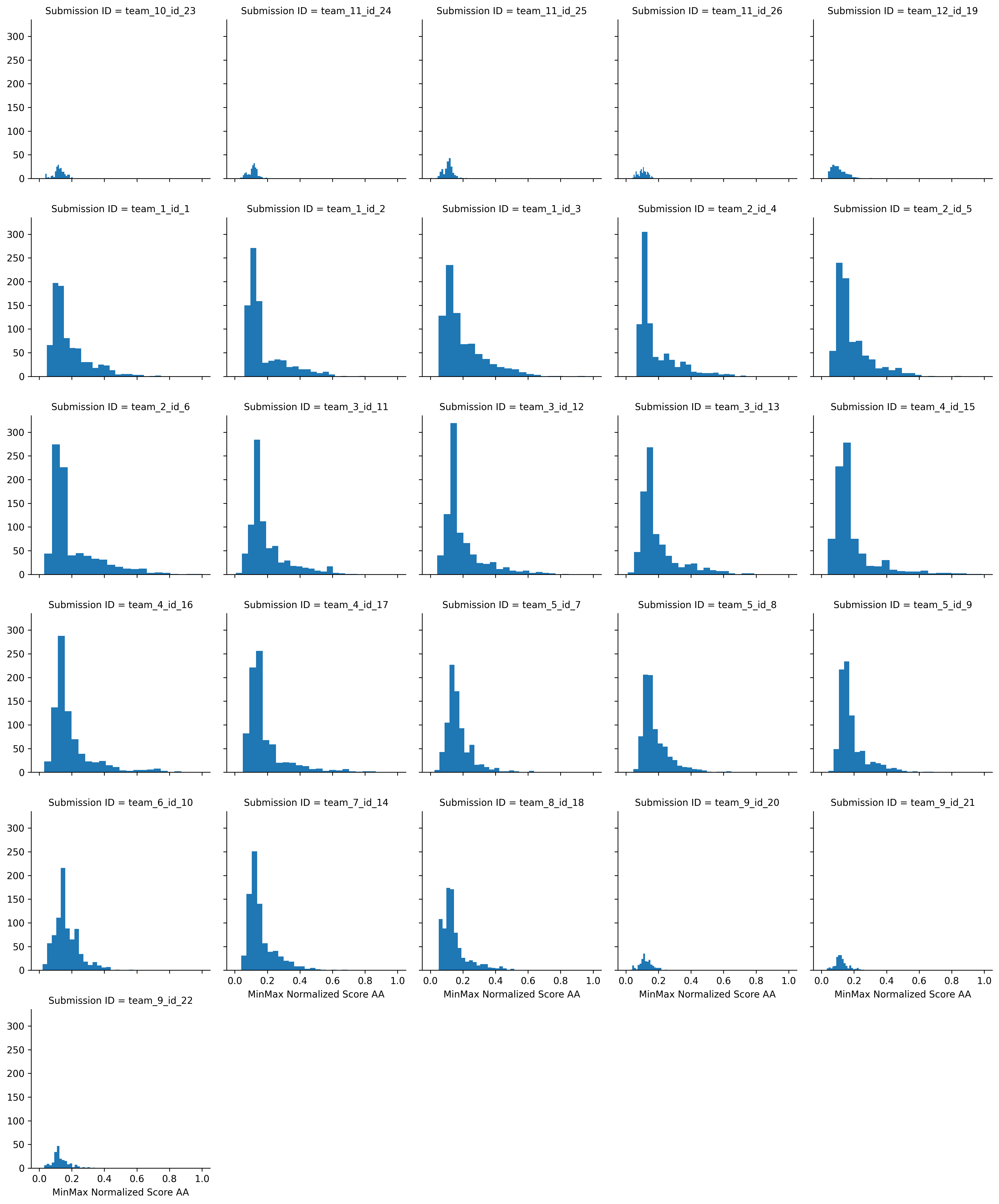}
    \caption{Min-Max normalised scores distributions for each submission.}
    \label{fig:scores_dist}
\end{figure*}

\begin{figure*}[ht]
    \centering
    \includegraphics[width=0.9\linewidth]{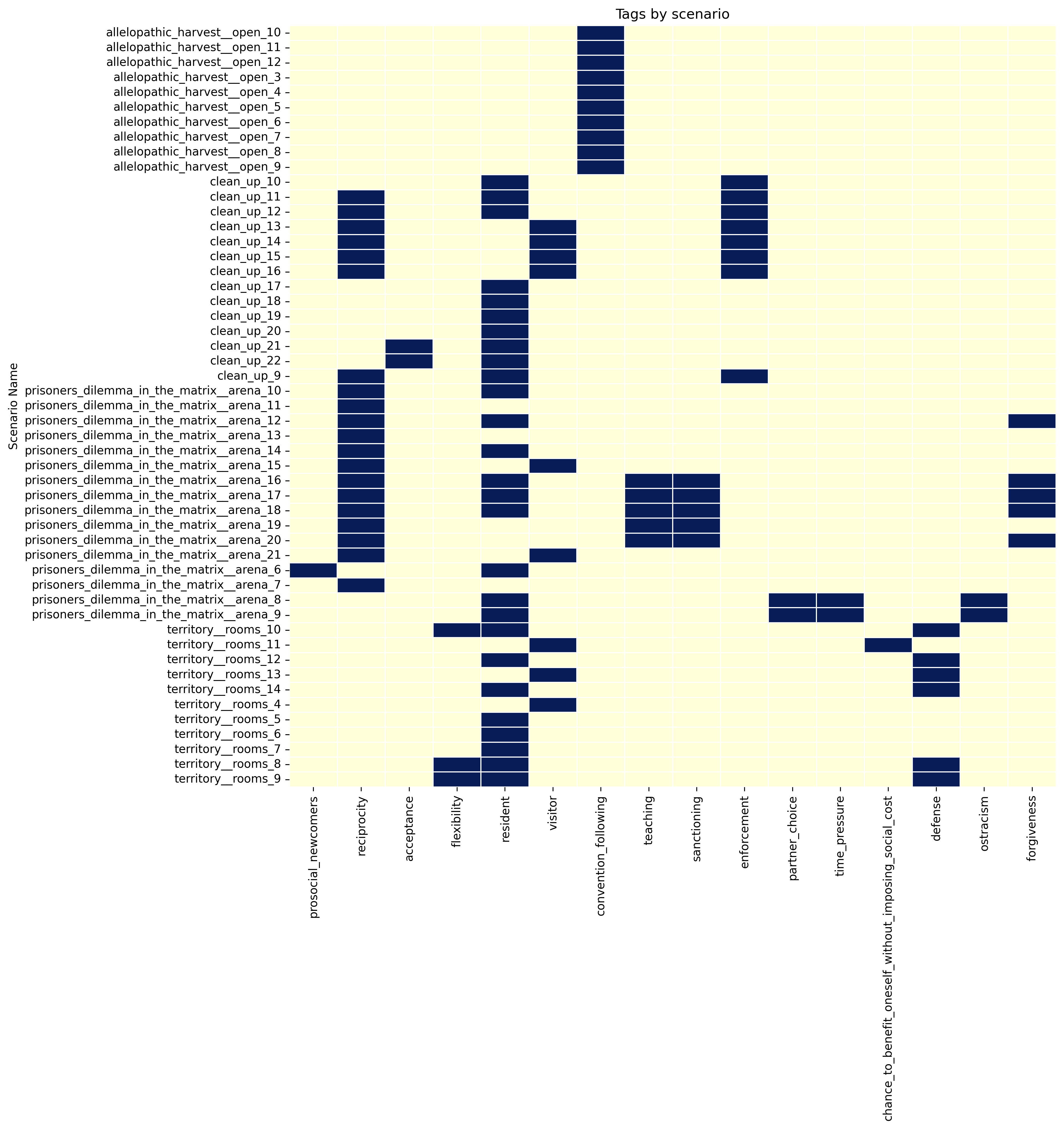}
    \caption{The plot shows whether each of the 16 demands is present or absent in each of the 51 scenarios. Blue indicates that the demand is present, while yellow indicates that it is absent.}
    \label{fig:tags_by_scenario}
\end{figure*}

\begin{figure*}[ht]
    \centering
    \includegraphics[width=0.9\linewidth]{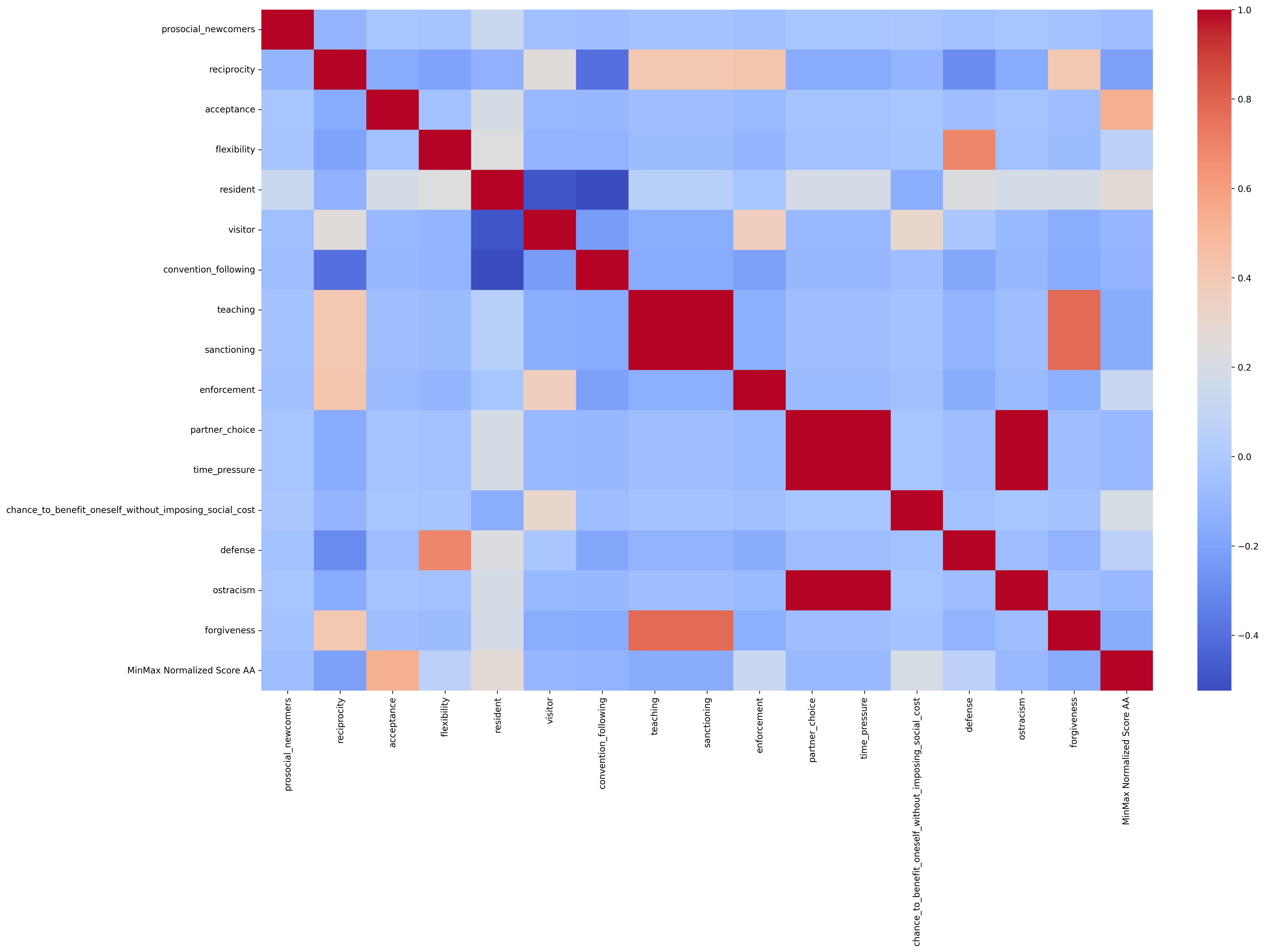}
    \caption{Correlation matrix for all 16 demands and the Min-Max normalised score.}
    \label{fig:corr_matrix}
\end{figure*}

\clearpage
\section{Measurement layouts}

\begin{figure*}[ht]
    \centering
    \includegraphics[width=0.95\linewidth]{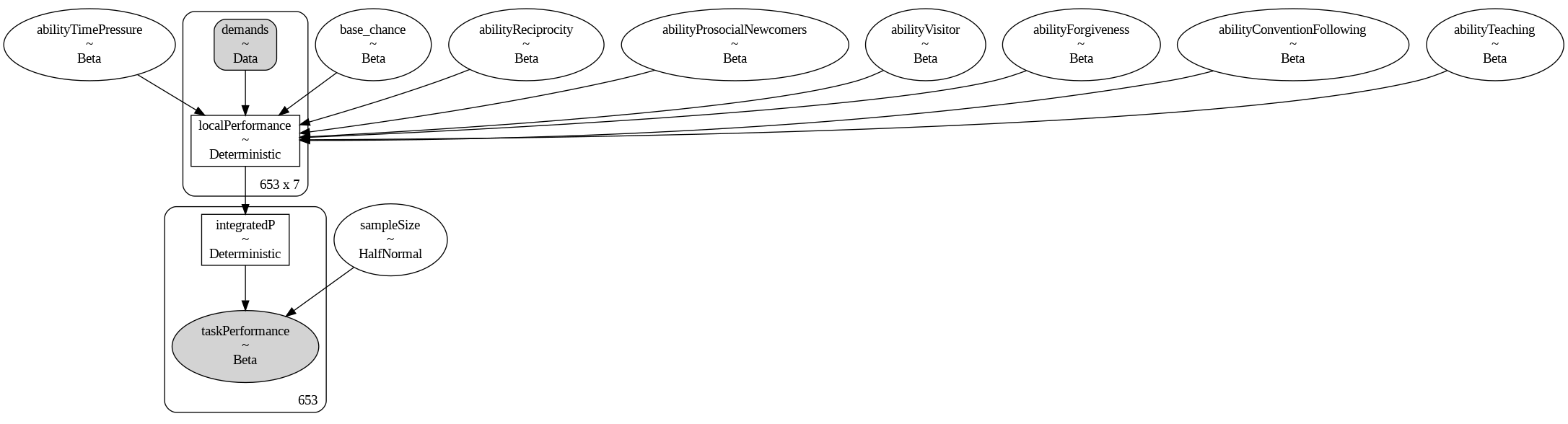}
    \caption{Graphical representation of the measurement layout used to model performance in the Melting Pot competition. Shaded variables indicate observed data. Each competition entry is assigned a separate measurement layout, where capabilities, base chance, and `sample size' are inferred using a Bayesian model.}
    \label{fig:mes_layout}
\end{figure*}

\begin{figure*}[ht]
    \centering
    
    \includegraphics[width=0.95\textwidth]{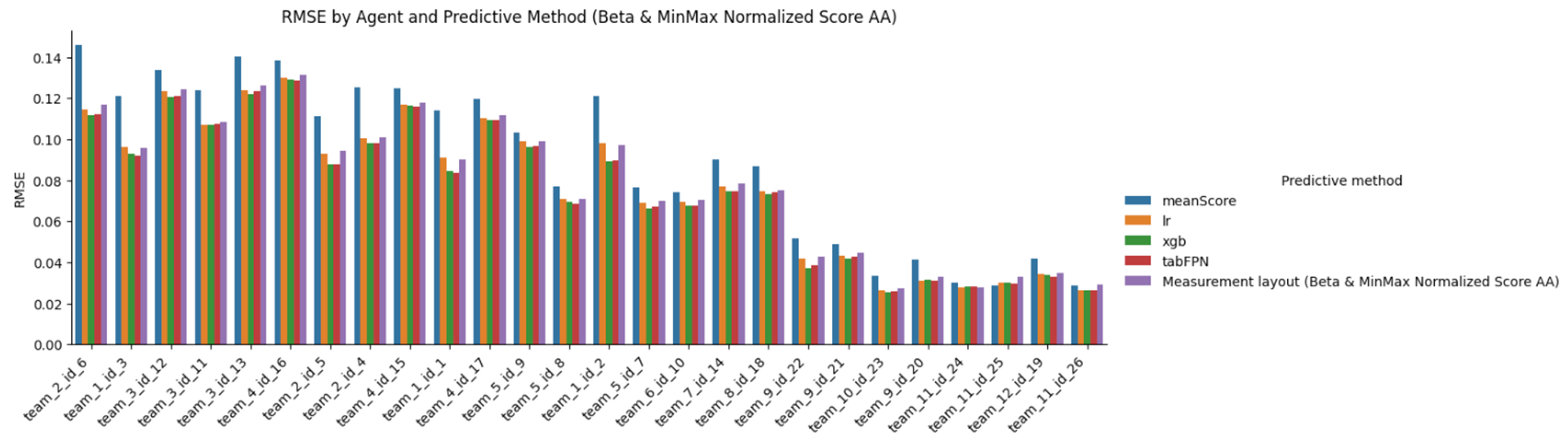}
    
    \includegraphics[width=0.95\textwidth]{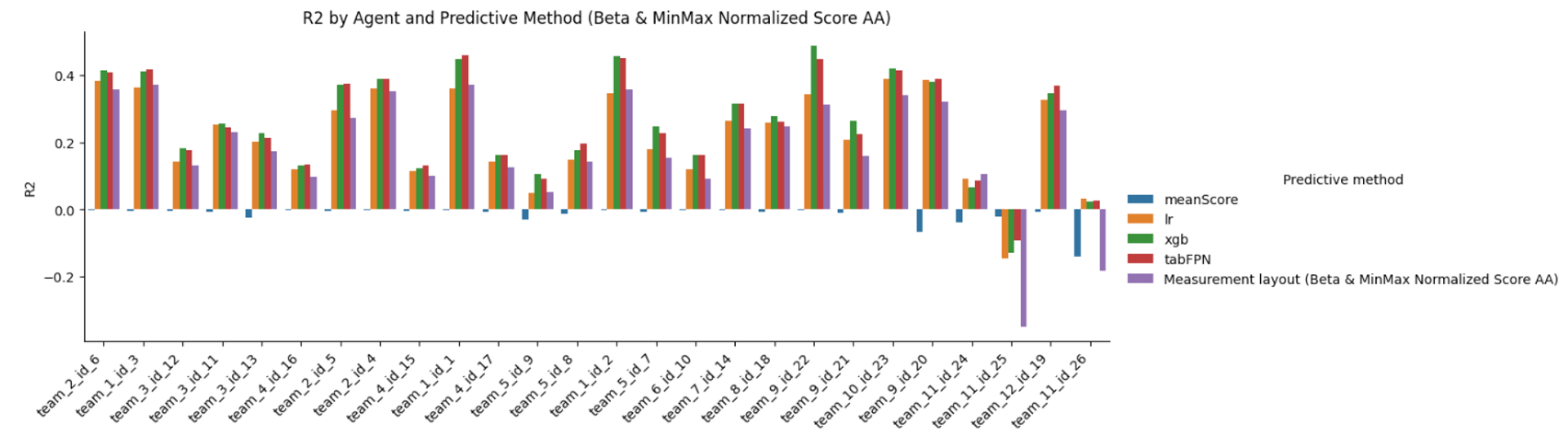}
    
    \caption{Predictive performance comparison of assessor models and measurement layouts. The top plot shows the RMSE values while the bottom plot presents the $R^2$ scores. Both assessor models and measurement layouts outperform the mean-score baseline for most submissions, with the exception of 
    team\_11\_id\_25.
    Note that the XGBoost regressor and tabPFN models, which can capture non-linear interactions, yield slightly higher predictive accuracy compared to linear regression and measurement layouts.}\label{fig:predictive_power}
\end{figure*}


\clearpage
\begin{longtable}{llcccccc}
    \caption{Full posterior summary statistics for all parameters and all competition entries. The table reports the posterior mean, highest density interval (HDI) at 3\% and 97\%, effective sample size (ESS) for bulk and tail, and the Gelman-Rubin convergence diagnostic ($\hat{R}$).} \label{tab:appendix_posterior_stats} \\
    \toprule
    \textbf{Submission ID} & \textbf{Parameter} & \textbf{Mean} & \textbf{HDI 3\%} & \textbf{HDI 97\%} & \textbf{ESS Bulk} & \textbf{ESS Tail} & $\bm{\hat{R}}$ \\
        \midrule
        team\_2\_id\_6  & abilityProsocialNewcomers  & 0.379  & 0.250  & 0.508  & 15697.167 & 10982.262 & 1.000 \\
        team\_2\_id\_6  & abilityVisitor            & 0.724  & 0.647  & 0.804  & 13628.207 & 9994.195  & 1.000 \\
        team\_2\_id\_6  & abilityTimePressure       & 0.389  & 0.295  & 0.483  & 15765.739 & 10620.590 & 1.000 \\
        team\_2\_id\_6  & abilityConventionFollowing & 0.351  & 0.310  & 0.395  & 13969.948 & 11110.248 & 1.001 \\
        team\_2\_id\_6  & abilityTeaching           & 0.742  & 0.579  & 0.913  & 8778.976  & 6205.097  & 1.001 \\
        team\_2\_id\_6  & abilityForgiveness        & 0.683  & 0.528  & 0.840  & 9596.650  & 7047.194  & 1.001 \\
        team\_2\_id\_6  & abilityReciprocity        & 0.538  & 0.487  & 0.589  & 12461.419 & 11055.615 & 1.000 \\
        team\_2\_id\_6  & base\_chance              & 0.853  & 0.846  & 0.859  & 13331.732 & 11870.175 & 1.001 \\
        team\_2\_id\_6  & sampleSize                & 10.826 & 9.756  & 11.937 & 15406.366 & 10818.480 & 1.000 \\
        team\_1\_id\_3  & abilityProsocialNewcomers  & 0.419  & 0.303  & 0.547  & 16361.931 & 10953.277 & 1.000 \\
        team\_1\_id\_3  & abilityVisitor            & 0.974  & 0.933  & 1.000  & 8366.691  & 5642.547  & 1.000 \\
        team\_1\_id\_3  & abilityTimePressure       & 0.394  & 0.310  & 0.479  & 16480.297 & 11314.092 & 1.000 \\
        team\_1\_id\_3  & abilityConventionFollowing & 0.570  & 0.517  & 0.622  & 14346.379 & 11919.436 & 1.000 \\
        team\_1\_id\_3  & abilityTeaching           & 0.767  & 0.611  & 0.927  & 8892.836  & 5461.931  & 1.000 \\
        team\_1\_id\_3  & abilityForgiveness        & 0.803  & 0.649  & 0.966  & 8581.964  & 5841.484  & 1.001 \\
        team\_1\_id\_3  & abilityReciprocity        & 0.416  & 0.379  & 0.452  & 12982.544 & 11616.137 & 1.001 \\
        team\_1\_id\_3  & base\_chance              & 0.837  & 0.831  & 0.842  & 13035.500 & 11475.855 & 1.000 \\
        team\_1\_id\_3  & sampleSize                & 15.230 & 13.691 & 16.779 & 16026.826 & 10737.616 & 1.000 \\
        team\_3\_id\_12 & abilityProsocialNewcomers  & 0.499  & 0.371  & 0.634  & 15954.154 & 10887.265 & 1.000 \\
        team\_3\_id\_12 & abilityVisitor            & 0.567  & 0.503  & 0.629  & 14977.278 & 11269.235 & 1.001 \\
        team\_3\_id\_12 & abilityTimePressure       & 0.488  & 0.385  & 0.596  & 15878.444 & 9480.959  & 1.000 \\
        team\_3\_id\_12 & abilityConventionFollowing & 0.602  & 0.546  & 0.655  & 13993.356 & 10940.649 & 1.000 \\
        team\_3\_id\_12 & abilityTeaching           & 0.719  & 0.587  & 0.865  & 11054.138 & 7260.792  & 1.000 \\
        team\_3\_id\_12 & abilityForgiveness        & 0.766  & 0.627  & 0.920  & 10634.426 & 6872.087  & 1.000 \\
        team\_3\_id\_12 & abilityReciprocity        & 0.612  & 0.556  & 0.666  & 11615.029 & 10004.089 & 1.000 \\
        team\_3\_id\_12 & base\_chance              & 0.827  & 0.821  & 0.834  & 12303.156 & 11862.654 & 1.000 \\
        team\_3\_id\_12 & sampleSize                & 15.668 & 14.052 & 17.245 & 16687.137 & 10887.484 & 1.000 \\
        team\_3\_id\_11 & abilityProsocialNewcomers  & 0.510  & 0.356  & 0.675  & 15902.320 & 10271.363 & 1.000 \\
        team\_3\_id\_11 & abilityVisitor            & 0.546  & 0.484  & 0.606  & 16356.287 & 11869.069 & 1.000 \\
        team\_3\_id\_11 & abilityTimePressure       & 0.516  & 0.401  & 0.630  & 16100.782 & 9894.277  & 1.000 \\
        team\_3\_id\_11 & abilityConventionFollowing & 0.613  & 0.554  & 0.669  & 14689.201 & 11236.494 & 1.000 \\
        team\_3\_id\_11 & abilityTeaching           & 0.730  & 0.585  & 0.869  & 11035.900 & 7371.841  & 1.000 \\
        team\_3\_id\_11 & abilityForgiveness        & 0.706  & 0.569  & 0.841  & 11815.645 & 8198.799  & 1.000 \\
        team\_3\_id\_11 & abilityReciprocity        & 0.664  & 0.606  & 0.725  & 12841.977 & 10793.681 & 1.000 \\
        team\_3\_id\_11 & base\_chance              & 0.824  & 0.818  & 0.830  & 13706.311 & 10805.159 & 1.000 \\
        team\_3\_id\_11 & sampleSize                & 15.314 & 13.806 & 16.851 & 18946.486 & 11798.402 & 1.001 \\
        team\_3\_id\_13 & abilityProsocialNewcomers  & 0.491  & 0.355  & 0.629  & 15854.740 & 9788.156  & 1.000 \\
        team\_3\_id\_13 & abilityVisitor            & 0.501  & 0.444  & 0.560  & 14484.315 & 11516.117 & 1.000 \\
        team\_3\_id\_13 & abilityTimePressure       & 0.524  & 0.425  & 0.623  & 15097.307 & 9887.652  & 1.000 \\
        team\_3\_id\_13 & abilityConventionFollowing & 0.633  & 0.576  & 0.690  & 13123.905 & 9635.083  & 1.000 \\
        team\_3\_id\_13 & abilityTeaching           & 0.696  & 0.565  & 0.837  & 10570.819 & 8765.955  & 1.001 \\
        team\_3\_id\_13 & abilityForgiveness        & 0.706  & 0.565  & 0.847  & 10220.876 & 6836.464  & 1.001 \\
        team\_3\_id\_13 & abilityReciprocity        & 0.682  & 0.624  & 0.742  & 12828.049 & 11918.486 & 1.000 \\
        team\_3\_id\_13 & base\_chance              & 0.822  & 0.816  & 0.828  & 11629.845 & 11276.493 & 1.000 \\
        team\_3\_id\_13 & sampleSize                & 16.240 & 14.626 & 17.869 & 15135.666 & 11609.856 & 1.000 \\
        team\_4\_id\_16 & abilityProsocialNewcomers  & 0.519  & 0.362  & 0.669  & 15842.305 & 9498.116  & 1.000 \\
        team\_4\_id\_16 & abilityVisitor            & 0.619  & 0.547  & 0.693  & 13159.968 & 10815.409 & 1.000 \\
        team\_4\_id\_16 & abilityTimePressure       & 0.551  & 0.431  & 0.672  & 14983.897 & 10680.727 & 1.001 \\
        team\_4\_id\_16 & abilityConventionFollowing & 0.653  & 0.587  & 0.722  & 14146.192 & 11298.987 & 1.000 \\
        team\_4\_id\_16 & abilityTeaching           & 0.709  & 0.567  & 0.859  & 10952.770 & 6898.595  & 1.000 \\
        team\_4\_id\_16 & abilityForgiveness        & 0.718  & 0.572  & 0.866  & 10949.583 & 7555.773  & 1.000 \\
        team\_4\_id\_16 & abilityReciprocity        & 0.712  & 0.645  & 0.780  & 11708.601 & 9412.500  & 1.000 \\
        team\_4\_id\_16 & base\_chance              & 0.819  & 0.812  & 0.826  & 12957.869 & 11727.520 & 1.000 \\
        team\_4\_id\_16 & sampleSize                & 11.882 & 10.599 & 13.031 & 15352.597 & 10574.253 & 1.000 \\
        team\_2\_id\_5  & abilityProsocialNewcomers  & 0.423  & 0.299  & 0.556  & 19046.461 & 11017.222 & 1.001 \\
        team\_2\_id\_5  & abilityVisitor            & 0.752  & 0.678  & 0.825  & 14324.413 & 10799.161 & 1.000 \\
        team\_2\_id\_5  & abilityTimePressure       & 0.395  & 0.319  & 0.471  & 16677.933 & 11256.621 & 1.000 \\
        team\_2\_id\_5  & abilityConventionFollowing & 0.564  & 0.517  & 0.611  & 15669.497 & 11495.211 & 1.000 \\
        team\_2\_id\_5  & abilityTeaching           & 0.808  & 0.664  & 0.961  & 9328.024  & 5211.757  & 1.001 \\
        team\_2\_id\_5  & abilityForgiveness        & 0.688  & 0.561  & 0.823  & 11020.836 & 8389.054  & 1.000 \\
        team\_2\_id\_5  & abilityReciprocity        & 0.497  & 0.454  & 0.540  & 12852.001 & 10754.192 & 1.000 \\
        team\_2\_id\_5  & base\_chance              & 0.828  & 0.823  & 0.833  & 13686.581 & 11383.569 & 1.000 \\
        team\_2\_id\_5  & sampleSize                & 20.395 & 18.368 & 22.481 & 16828.401 & 9916.925  & 1.001 \\
        team\_2\_id\_4  & abilityProsocialNewcomers  & 0.395  & 0.288  & 0.494  & 15651.587 & 10464.587 & 1.001 \\
        team\_2\_id\_4  & abilityVisitor            & 0.707  & 0.643  & 0.767  & 14240.154 & 10191.300 & 1.000 \\
        team\_2\_id\_4  & abilityTimePressure       & 0.397  & 0.318  & 0.479  & 15707.963 & 10574.875 & 1.000 \\
        team\_2\_id\_4  & abilityConventionFollowing & 0.367  & 0.331  & 0.404  & 15093.042 & 11232.845 & 1.000 \\
        team\_2\_id\_4  & abilityTeaching           & 0.761  & 0.627  & 0.898  & 9579.043  & 6635.528  & 1.001 \\
        team\_2\_id\_4  & abilityForgiveness        & 0.642  & 0.525  & 0.758  & 11370.851 & 10014.292 & 1.000 \\
        team\_2\_id\_4  & abilityReciprocity        & 0.554  & 0.514  & 0.599  & 13851.727 & 10583.129 & 1.000 \\
        team\_2\_id\_4  & base\_chance              & 0.834  & 0.829  & 0.840  & 12847.286 & 11698.157 & 1.000 \\
        team\_2\_id\_4  & sampleSize                & 19.936 & 17.893 & 21.952 & 14186.901 & 11079.777 & 1.000 \\
        team\_4\_id\_15 & abilityProsocialNewcomers  & 0.584  & 0.414  & 0.761  & 15317.012 & 8235.246  & 1.000 \\
        team\_4\_id\_15 & abilityVisitor            & 0.508  & 0.443  & 0.573  & 14483.282 & 10549.932 & 1.001 \\
        team\_4\_id\_15 & abilityTimePressure       & 0.629  & 0.494  & 0.770  & 14052.426 & 8733.930  & 1.000 \\
        team\_4\_id\_15 & abilityConventionFollowing & 0.700  & 0.624  & 0.775  & 12487.175 & 9949.673  & 1.000 \\
        team\_4\_id\_15 & abilityTeaching           & 0.679  & 0.531  & 0.831  & 12755.346 & 8209.939  & 1.000 \\
        team\_4\_id\_15 & abilityForgiveness        & 0.681  & 0.535  & 0.829  & 11964.054 & 7475.856  & 1.001 \\
        team\_4\_id\_15 & abilityReciprocity        & 0.820  & 0.744  & 0.905  & 11400.131 & 8117.964  & 1.000 \\
        team\_4\_id\_15 & base\_chance              & 0.811  & 0.803  & 0.819  & 10123.751 & 10382.835 & 1.001 \\
        team\_4\_id\_15 & sampleSize                & 10.500 & 9.419  & 11.555 & 15338.808 & 10911.070 & 1.000 \\
        team\_1\_id\_1  & abilityProsocialNewcomers  & 0.413  & 0.298  & 0.536  & 17243.707 & 11315.390 & 1.000 \\
        team\_1\_id\_1  & abilityVisitor            & 0.981  & 0.950  & 1.000  & 11725.210 & 6551.323  & 1.000 \\
        team\_1\_id\_1  & abilityTimePressure       & 0.410  & 0.329  & 0.490  & 17592.423 & 10824.634 & 1.000 \\
        team\_1\_id\_1  & abilityConventionFollowing & 0.613  & 0.563  & 0.662  & 15025.013 & 11473.691 & 1.000 \\
        team\_1\_id\_1  & abilityTeaching           & 0.768  & 0.623  & 0.915  & 10052.907 & 6318.440  & 1.000 \\
        team\_1\_id\_1  & abilityForgiveness        & 0.791  & 0.649  & 0.947  & 8872.270  & 5157.133  & 1.000 \\
        team\_1\_id\_1  & abilityReciprocity        & 0.421  & 0.387  & 0.455  & 13574.997 & 12079.213 & 1.000 \\
        team\_1\_id\_1  & base\_chance              & 0.824  & 0.818  & 0.829  & 13333.667 & 11903.165 & 1.000 \\
        team\_1\_id\_1  & sampleSize                & 20.402 & 18.229 & 22.340 & 16641.534 & 10833.029 & 1.000 \\
        team\_4\_id\_17 & abilityProsocialNewcomers  & 0.570  & 0.382  & 0.771  & 16333.730 & 8649.721  & 1.001 \\
        team\_4\_id\_17 & abilityVisitor            & 0.518  & 0.457  & 0.582  & 14103.592 & 11203.386 & 1.000 \\
        team\_4\_id\_17 & abilityTimePressure       & 0.591  & 0.462  & 0.724  & 14234.057 & 9683.552  & 1.000 \\
        team\_4\_id\_17 & abilityConventionFollowing & 0.698  & 0.624  & 0.769  & 12407.590 & 9772.103  & 1.000 \\
        team\_4\_id\_17 & abilityTeaching           & 0.633  & 0.507  & 0.765  & 11792.877 & 8925.113  & 1.000 \\
        team\_4\_id\_17 & abilityForgiveness        & 0.687  & 0.551  & 0.831  & 10455.895 & 7205.825  & 1.000 \\
        team\_4\_id\_17 & abilityReciprocity        & 0.834  & 0.756  & 0.911  & 10197.260 & 7990.684  & 1.000 \\
        team\_4\_id\_17 & base\_chance              & 0.806  & 0.798  & 0.813  & 10957.438 & 11302.147 & 1.000 \\
        team\_4\_id\_17 & sampleSize                & 12.495 & 11.202 & 13.778 & 15808.188 & 11397.013 & 1.000 \\
        team\_5\_id\_9  & abilityProsocialNewcomers  & 0.548  & 0.427  & 0.674  & 14477.033 & 8895.705  & 1.000 \\
        team\_5\_id\_9  & abilityVisitor            & 0.635  & 0.581  & 0.692  & 13481.370 & 10729.674 & 1.000 \\
        team\_5\_id\_9  & abilityTimePressure       & 0.541  & 0.448  & 0.628  & 15536.857 & 9965.804  & 1.001 \\
        team\_5\_id\_9  & abilityConventionFollowing & 0.678  & 0.629  & 0.729  & 13336.752 & 11174.523 & 1.000 \\
        team\_5\_id\_9  & abilityTeaching           & 0.732  & 0.620  & 0.844  & 10887.689 & 7803.281  & 1.001 \\
        team\_5\_id\_9  & abilityForgiveness        & 0.679  & 0.575  & 0.779  & 12024.946 & 10018.589 & 1.000 \\
        team\_5\_id\_9  & abilityReciprocity        & 0.705  & 0.656  & 0.757  & 11825.039 & 10363.827 & 1.001 \\
        team\_5\_id\_9  & base\_chance              & 0.803  & 0.797  & 0.808  & 12298.243 & 11538.941 & 1.000 \\
        team\_5\_id\_9  & sampleSize                & 27.929 & 25.166 & 30.679 & 15033.376 & 11311.357 & 1.000 \\
        team\_5\_id\_8  & abilityProsocialNewcomers  & 0.480  & 0.365  & 0.594  & 16451.630 & 10817.221 & 1.000 \\
        team\_5\_id\_8  & abilityVisitor            & 0.676  & 0.618  & 0.733  & 15962.458 & 11424.793 & 1.000 \\
        team\_5\_id\_8  & abilityTimePressure       & 0.453  & 0.374  & 0.534  & 15585.492 & 10828.598 & 1.000 \\
        team\_5\_id\_8  & abilityConventionFollowing & 0.674  & 0.623  & 0.725  & 14194.413 & 10917.423 & 1.000 \\
        team\_5\_id\_8  & abilityTeaching           & 0.704  & 0.587  & 0.828  & 12134.799 & 8775.740  & 1.000 \\
        team\_5\_id\_8  & abilityForgiveness        & 0.727  & 0.603  & 0.854  & 11977.119 & 8668.898  & 1.000 \\
        team\_5\_id\_8  & abilityReciprocity        & 0.633  & 0.589  & 0.681  & 12720.377 & 10823.879 & 1.000 \\
        team\_5\_id\_8  & base\_chance              & 0.808  & 0.802  & 0.813  & 12200.642 & 11290.650 & 1.000 \\
        team\_5\_id\_8  & sampleSize                & 27.277 & 24.680 & 30.121 & 15346.459 & 11017.762 & 1.000 \\
        team\_1\_id\_2  & abilityProsocialNewcomers  & 0.421  & 0.302  & 0.536  & 17352.072 & 10234.988 & 1.000 \\
        team\_1\_id\_2  & abilityVisitor            & 0.987  & 0.965  & 1.000  & 13048.606 & 8326.609  & 1.000 \\
        team\_1\_id\_2  & abilityTimePressure       & 0.409  & 0.330  & 0.490  & 16323.606 & 10594.598 & 1.000 \\
        team\_1\_id\_2  & abilityConventionFollowing & 0.438  & 0.398  & 0.479  & 14579.104 & 11206.331 & 1.000 \\
        team\_1\_id\_2  & abilityTeaching           & 0.807  & 0.654  & 0.965  & 8127.996  & 6000.194  & 1.000 \\
        team\_1\_id\_2  & abilityForgiveness        & 0.794  & 0.646  & 0.949  & 6951.310  & 3497.004  & 1.001 \\
        team\_1\_id\_2  & abilityReciprocity        & 0.408  & 0.376  & 0.440  & 11569.019 & 10735.436 & 1.000 \\
        team\_1\_id\_2  & base\_chance              & 0.823  & 0.818  & 0.828  & 12184.492 & 11427.784 & 1.000 \\
        team\_1\_id\_2  & sampleSize                & 21.226 & 19.107 & 23.382 & 15374.191 & 11240.901 & 1.000 \\
        team\_5\_id\_7  & abilityProsocialNewcomers  & 0.536  & 0.428  & 0.647  & 17507.938 & 10927.937 & 1.000 \\
        team\_5\_id\_7  & abilityVisitor            & 0.612  & 0.559  & 0.669  & 13703.911 & 10689.297 & 1.000 \\
        team\_5\_id\_7  & abilityTimePressure       & 0.514  & 0.432  & 0.600  & 15074.062 & 10528.661 & 1.000 \\
        team\_5\_id\_7  & abilityConventionFollowing & 0.693  & 0.644  & 0.742  & 12451.011 & 10453.552 & 1.000 \\
        team\_5\_id\_7  & abilityTeaching           & 0.748  & 0.633  & 0.870  & 12151.090 & 8637.760  & 1.000 \\
        team\_5\_id\_7  & abilityForgiveness        & 0.795  & 0.669  & 0.917  & 10511.084 & 6295.438  & 1.000 \\
        team\_5\_id\_7  & abilityReciprocity        & 0.590  & 0.546  & 0.636  & 11719.342 & 10633.456 & 1.000 \\
        team\_5\_id\_7  & base\_chance              & 0.803  & 0.798  & 0.808  & 11580.302 & 11288.044 & 1.000 \\
        team\_5\_id\_7  & sampleSize                & 30.358 & 27.418 & 33.440 & 18064.756 & 10525.454 & 1.000 \\
        team\_6\_id\_10 & abilityProsocialNewcomers  & 0.574  & 0.451  & 0.708  & 15805.018 & 9722.371  & 1.000 \\
        team\_6\_id\_10 & abilityVisitor            & 0.506  & 0.456  & 0.560  & 15178.938 & 10497.083 & 1.000 \\
        team\_6\_id\_10 & abilityTimePressure       & 0.677  & 0.573  & 0.781  & 15415.116 & 10232.110 & 1.000 \\
        team\_6\_id\_10 & abilityConventionFollowing & 0.733  & 0.680  & 0.790  & 14596.488 & 11236.767 & 1.000 \\
        team\_6\_id\_10 & abilityTeaching           & 0.775  & 0.644  & 0.899  & 9801.914  & 7208.946  & 1.000 \\
        team\_6\_id\_10 & abilityForgiveness        & 0.743  & 0.626  & 0.870  & 9587.141  & 6992.855  & 1.000 \\
        team\_6\_id\_10 & abilityReciprocity        & 0.672  & 0.619  & 0.727  & 13376.832 & 11230.742 & 1.000 \\
        team\_6\_id\_10 & base\_chance              & 0.795  & 0.789  & 0.800  & 12375.390 & 11374.892 & 1.000 \\
        team\_6\_id\_10 & sampleSize                & 27.029 & 24.412 & 29.706 & 17469.573 & 11515.122 & 1.001 \\
        team\_7\_id\_14 & abilityProsocialNewcomers  & 0.541  & 0.422  & 0.652  & 18996.657 & 11318.662 & 1.000 \\
        team\_7\_id\_14 & abilityVisitor            & 0.628  & 0.568  & 0.690  & 15919.142 & 10447.999 & 1.001 \\
        team\_7\_id\_14 & abilityTimePressure       & 0.567  & 0.478  & 0.663  & 16698.115 & 10621.313 & 1.000 \\
        team\_7\_id\_14 & abilityConventionFollowing & 0.414  & 0.377  & 0.451  & 14533.700 & 11390.850 & 1.000 \\
        team\_7\_id\_14 & abilityTeaching           & 0.755  & 0.631  & 0.881  & 10617.996 & 7606.926  & 1.001 \\
        team\_7\_id\_14 & abilityForgiveness        & 0.759  & 0.639  & 0.887  & 10663.812 & 6768.806  & 1.001 \\
        team\_7\_id\_14 & abilityReciprocity        & 0.614  & 0.567  & 0.664  & 13588.484 & 11094.694 & 1.000 \\
        team\_7\_id\_14 & base\_chance              & 0.803  & 0.798  & 0.808  & 13371.729 & 12158.265 & 1.000 \\
        team\_7\_id\_14 & sampleSize                & 28.316 & 25.614 & 31.122 & 17077.802 & 11913.959 & 1.000 \\
        team\_8\_id\_18 & abilityProsocialNewcomers  & 0.595  & 0.459  & 0.725  & 14714.378 & 8464.834  & 1.001 \\
        team\_8\_id\_18 & abilityVisitor            & 0.681  & 0.627  & 0.739  & 14728.773 & 10527.711 & 1.000 \\
        team\_8\_id\_18 & abilityTimePressure       & 0.549  & 0.460  & 0.644  & 16429.678 & 10698.240 & 1.000 \\
        team\_8\_id\_18 & abilityConventionFollowing & 0.378  & 0.342  & 0.416  & 15798.249 & 11572.567 & 1.000 \\
        team\_8\_id\_18 & abilityTeaching           & 0.599  & 0.504  & 0.698  & 11853.133 & 10685.729 & 1.000 \\
        team\_8\_id\_18 & abilityForgiveness        & 0.765  & 0.644  & 0.891  & 10623.699 & 6489.129  & 1.000 \\
        team\_8\_id\_18 & abilityReciprocity        & 0.664  & 0.617  & 0.711  & 13605.909 & 11746.296 & 1.000 \\
        team\_8\_id\_18 & base\_chance              & 0.791  & 0.785  & 0.796  & 12283.994 & 11944.945 & 1.000 \\
        team\_8\_id\_18 & sampleSize                & 29.737 & 26.801 & 32.685 & 16093.607 & 10476.502 & 1.000 \\
        team\_9\_id\_22 & abilityProsocialNewcomers  & 0.592  & 0.295  & 0.912  & 13791.271 & 6818.620  & 1.001 \\
        team\_9\_id\_22 & abilityVisitor            & 0.752  & 0.611  & 0.897  & 8251.814  & 5680.622  & 1.001 \\
        team\_9\_id\_22 & abilityTimePressure       & 0.575  & 0.381  & 0.775  & 14223.384 & 8215.767  & 1.001 \\
        team\_9\_id\_22 & abilityConventionFollowing & 0.601  & 0.500  & 0.699  & 11958.526 & 9510.856  & 1.000 \\
        team\_9\_id\_22 & abilityTeaching           & 0.836  & 0.664  & 1.000  & 9285.657  & 5638.872  & 1.000 \\
        team\_9\_id\_22 & abilityForgiveness        & 0.817  & 0.648  & 1.000  & 9185.038  & 6899.815  & 1.000 \\
        team\_9\_id\_22 & abilityReciprocity        & 0.475  & 0.393  & 0.555  & 9374.468  & 8468.892  & 1.001 \\
        team\_9\_id\_22 & base\_chance              & 0.780  & 0.769  & 0.790  & 10068.694 & 10295.242 & 1.000 \\
        team\_9\_id\_22 & sampleSize                & 31.470 & 26.053 & 36.748 & 13926.668 & 10908.318 & 1.000 \\
        team\_9\_id\_21 & abilityProsocialNewcomers  & 0.610  & 0.358  & 0.887  & 15193.607 & 6890.213  & 1.001 \\
        team\_9\_id\_21 & abilityVisitor            & 0.572  & 0.460  & 0.684  & 13202.985 & 9940.412  & 1.000 \\
        team\_9\_id\_21 & abilityTimePressure       & 0.624  & 0.410  & 0.857  & 11230.258 & 6090.162  & 1.000 \\
        team\_9\_id\_21 & abilityConventionFollowing & 0.627  & 0.526  & 0.727  & 11814.083 & 10099.697 & 1.000 \\
        team\_9\_id\_21 & abilityTeaching           & 0.738  & 0.546  & 0.953  & 9651.357  & 5992.842  & 1.001 \\
        team\_9\_id\_21 & abilityForgiveness        & 0.774  & 0.584  & 0.982  & 10435.661 & 6684.995  & 1.000 \\
        team\_9\_id\_21 & abilityReciprocity        & 0.600  & 0.502  & 0.692  & 10314.895 & 8894.649  & 1.000 \\
        team\_9\_id\_21 & base\_chance              & 0.769  & 0.758  & 0.779  & 10155.840 & 9883.795  & 1.000 \\
        team\_9\_id\_21 & sampleSize                & 35.374 & 29.385 & 41.159 & 14893.482 & 10229.818 & 1.000 \\
        team\_10\_id\_23 & abilityProsocialNewcomers  & 0.737  & 0.513  & 0.979  & 9916.809  & 5241.002  & 1.000 \\
        team\_10\_id\_23 & abilityVisitor            & 0.537  & 0.434  & 0.641  & 12215.029 & 10455.746 & 1.000 \\
        team\_10\_id\_23 & abilityTimePressure       & 0.656  & 0.427  & 0.886  & 11696.790 & 5637.696  & 1.001 \\
        team\_10\_id\_23 & abilityConventionFollowing & 0.733  & 0.621  & 0.851  & 10839.900 & 8611.850  & 1.001 \\
        team\_10\_id\_23 & abilityTeaching           & 0.787  & 0.613  & 1.000  & 6433.369  & 3989.326  & 1.000 \\
        team\_10\_id\_23 & abilityForgiveness        & 0.764  & 0.586  & 0.989  & 6079.437  & 4334.794  & 1.001 \\
        team\_10\_id\_23 & abilityReciprocity        & 0.651  & 0.551  & 0.757  & 8953.218  & 9033.628  & 1.000 \\
        team\_10\_id\_23 & base\_chance              & 0.760  & 0.749  & 0.771  & 9676.885  & 11312.462 & 1.000 \\
        team\_10\_id\_23 & sampleSize                & 36.251 & 30.471 & 42.005 & 15908.751 & 11310.362 & 1.000 \\
        team\_9\_id\_20  & abilityProsocialNewcomers  & 0.703  & 0.475  & 0.968  & 10393.024 & 4958.908  & 1.001 \\
        team\_9\_id\_20  & abilityVisitor            & 0.608  & 0.484  & 0.726  & 13518.498 & 9989.146  & 1.000 \\
        team\_9\_id\_20  & abilityTimePressure       & 0.638  & 0.429  & 0.842  & 13735.806 & 8153.861  & 1.000 \\
        team\_9\_id\_20  & abilityConventionFollowing & 0.644  & 0.539  & 0.749  & 11583.881 & 8691.827  & 1.000 \\
        team\_9\_id\_20  & abilityTeaching           & 0.727  & 0.519  & 0.960  & 9991.555  & 5508.111  & 1.000 \\
        team\_9\_id\_20  & abilityForgiveness        & 0.749  & 0.540  & 0.969  & 10287.297 & 6757.767  & 1.000 \\
        team\_9\_id\_20  & abilityReciprocity        & 0.588  & 0.490  & 0.686  & 11555.595 & 10084.761 & 1.000 \\
        team\_9\_id\_20  & base\_chance              & 0.768  & 0.757  & 0.779  & 10159.243 & 10834.109 & 1.000 \\
        team\_9\_id\_20  & sampleSize                & 34.318 & 28.391 & 39.779 & 15920.981 & 10891.215 & 1.000 \\
        team\_11\_id\_24 & abilityProsocialNewcomers  & 0.767  & 0.533  & 1.000  & 10417.328 & 6153.939  & 1.000 \\
        team\_11\_id\_24 & abilityVisitor            & 0.599  & 0.487  & 0.717  & 14239.165 & 10393.165 & 1.000 \\
        team\_11\_id\_24 & abilityTimePressure       & 0.822  & 0.649  & 1.000  & 8531.254  & 5391.709  & 1.000 \\
        team\_11\_id\_24 & abilityConventionFollowing & 0.638  & 0.530  & 0.751  & 11699.362 & 9373.529  & 1.000 \\
        team\_11\_id\_24 & abilityTeaching           & 0.860  & 0.702  & 1.000  & 9065.428  & 5771.847  & 1.000 \\
        team\_11\_id\_24 & abilityForgiveness        & 0.768  & 0.587  & 0.966  & 9165.625  & 5161.234  & 1.000 \\
        team\_11\_id\_24 & abilityReciprocity        & 0.668  & 0.567  & 0.775  & 11280.424 & 10274.945 & 1.000 \\
        team\_11\_id\_24 & base\_chance              & 0.743  & 0.732  & 0.754  & 10384.826 & 11595.247 & 1.000 \\
        team\_11\_id\_24 & sampleSize                & 37.055 & 31.141 & 43.094 & 13605.939 & 10385.776 & 1.000 \\
        team\_11\_id\_25 & abilityProsocialNewcomers  & 0.714  & 0.432  & 1.000  & 10454.116 & 5915.658  & 1.001 \\
        team\_11\_id\_25 & abilityVisitor            & 0.498  & 0.392  & 0.607  & 14087.274 & 8994.261  & 1.000 \\
        team\_11\_id\_25 & abilityTimePressure       & 0.757  & 0.552  & 0.971  & 9252.433  & 5012.188  & 1.000 \\
        team\_11\_id\_25 & abilityConventionFollowing & 0.631  & 0.517  & 0.750  & 13967.014 & 10548.741 & 1.000 \\
        team\_11\_id\_25 & abilityTeaching           & 0.782  & 0.607  & 0.993  & 8565.744  & 5883.564  & 1.001 \\
        team\_11\_id\_25 & abilityForgiveness        & 0.740  & 0.552  & 0.952  & 7764.295  & 5175.775  & 1.001 \\
        team\_11\_id\_25 & abilityReciprocity        & 0.744  & 0.621  & 0.862  & 10831.915 & 7644.696  & 1.000 \\
        team\_11\_id\_25 & base\_chance              & 0.743  & 0.731  & 0.755  & 10939.227 & 11412.498 & 1.001 \\
        team\_11\_id\_25 & sampleSize                & 32.892 & 27.293 & 38.614 & 14645.017 & 10644.047 & 1.001 \\
        team\_12\_id\_19 & abilityProsocialNewcomers  & 0.584  & 0.310  & 0.856  & 12892.308 & 7095.603  & 1.000 \\
        team\_12\_id\_19 & abilityVisitor            & 0.526  & 0.419  & 0.632  & 14430.535 & 9880.540  & 1.000 \\
        team\_12\_id\_19 & abilityTimePressure       & 0.582  & 0.390  & 0.771  & 13587.588 & 8077.461  & 1.001 \\
        team\_12\_id\_19 & abilityConventionFollowing & 0.447  & 0.364  & 0.533  & 13465.905 & 11731.943 & 1.000 \\
        team\_12\_id\_19 & abilityTeaching           & 0.705  & 0.496  & 0.944  & 10650.146 & 6253.214  & 1.000 \\
        team\_12\_id\_19 & abilityForgiveness        & 0.767  & 0.578  & 0.993  & 8968.174  & 5538.929  & 1.000 \\
        team\_12\_id\_19 & abilityReciprocity        & 0.564  & 0.474  & 0.655  & 11711.390 & 9139.178  & 1.000 \\
        team\_12\_id\_19 & base\_chance              & 0.764  & 0.753  & 0.776  & 10637.953 & 10522.990 & 1.000 \\
        team\_12\_id\_19 & sampleSize                & 33.922 & 28.144 & 39.588 & 16328.691 & 10833.556 & 1.000 \\
        team\_11\_id\_26 & abilityProsocialNewcomers  & 0.781  & 0.549  & 1.000  & 11859.506 & 5980.572  & 1.000 \\
        team\_11\_id\_26 & abilityVisitor            & 0.627  & 0.506  & 0.755  & 12578.044 & 8822.489  & 1.000 \\
        team\_11\_id\_26 & abilityTimePressure       & 0.845  & 0.688  & 1.000  & 8498.034  & 4814.647  & 1.000 \\
        team\_11\_id\_26 & abilityConventionFollowing & 0.579  & 0.478  & 0.684  & 11688.582 & 10157.464 & 1.000 \\
        team\_11\_id\_26 & abilityTeaching           & 0.822  & 0.655  & 1.000  & 8439.728  & 5648.368  & 1.000 \\
        team\_11\_id\_26 & abilityForgiveness        & 0.782  & 0.613  & 0.994  & 8523.905  & 5340.397  & 1.000 \\
        team\_11\_id\_26 & abilityReciprocity        & 0.653  & 0.556  & 0.756  & 9811.378  & 9309.728  & 1.000 \\
        team\_11\_id\_26 & base\_chance              & 0.742  & 0.731  & 0.753  & 9515.819  & 10665.776 & 1.000 \\
        team\_11\_id\_26 & sampleSize                & 37.803 & 31.786 & 44.004 & 16228.747 & 10140.086 & 1.000 \\
        \bottomrule
\end{longtable}

\begin{figure*}[ht]
\centering

\includegraphics[width=0.95\textwidth]{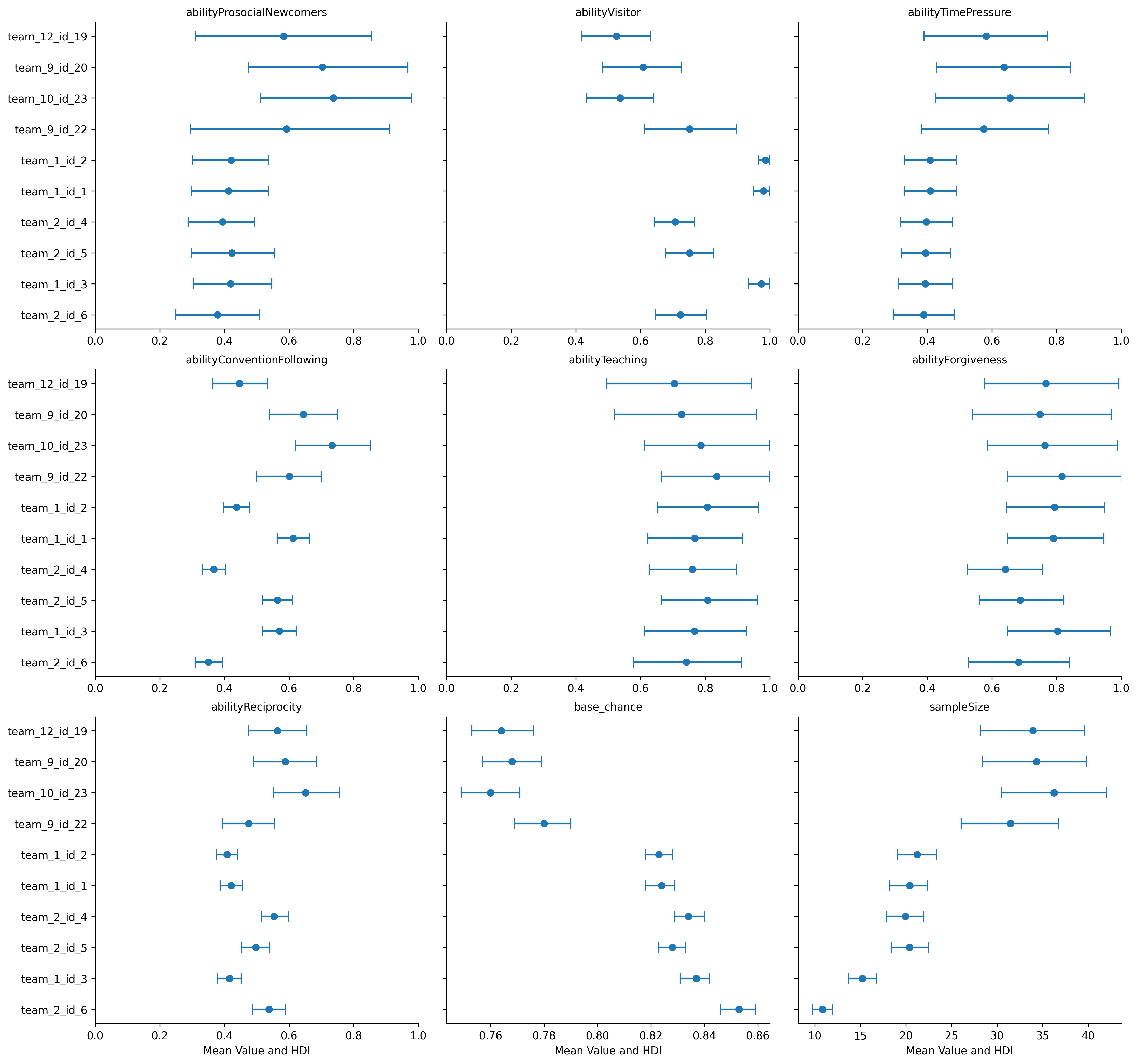}

\caption{Average posterior estimates for each ability, along with their corresponding highest density interval (HDI) bounds.}\label{fig:posteriors_hdi}
\end{figure*}